\documentclass[final,3p,times]{elsarticle}

\usepackage{url}

\usepackage{adjustbox}
\usepackage[utf8]{inputenc}

\usepackage{graphicx}%
\usepackage{multirow}%
\usepackage{amsmath,amssymb,amsfonts}%
\usepackage{amsthm}%
\usepackage{mathrsfs}%
\usepackage[title]{appendix}%

\usepackage[dvipsnames]{xcolor}
\usepackage{xcolor}%
\usepackage{textcomp}%
\usepackage{manyfoot}%
\usepackage{booktabs}%
\usepackage{algorithm}%
\usepackage{algorithmicx}%
\usepackage{algpseudocode}%
\usepackage{listings}%
\usepackage{natbib}%
\usepackage{tabularx}
\usepackage{tabulary}
\usepackage{threeparttable} 
\usepackage{comment}

\usepackage{hyperref}

\usepackage{times}
\usepackage{epsfig}
\usepackage{multicol}
\usepackage{makecell}
\usepackage{enumitem}
\usepackage{array}
\usepackage{siunitx}
\usepackage{mathtools}
\usepackage{nomencl}
\usepackage{soul}
\usepackage[usestackEOL]{stackengine}
\usepackage{soul}

\usepackage{listings}
\definecolor{codegreen}{rgb}{0,0.6,0}
\definecolor{codegray}{rgb}{0.5,0.5,0.5}
\definecolor{codepurple}{rgb}{0.58,0,0.82}
\definecolor{backcolour}{rgb}{0.95,0.95,0.92}

\newcommand{\thickbar}[1]{\mathbf{\bar{\text{$#1$}}}}

\lstdefinestyle{mystyle}{
    backgroundcolor=\color{backcolour},   
    commentstyle=\color{codegreen},
    keywordstyle=\color{magenta},
    numberstyle=\tiny\color{codegray},
    stringstyle=\color{codepurple},
    basicstyle=\ttfamily\footnotesize,
    breakatwhitespace=false,         
    breaklines=true,                 
    captionpos=b,                    
    keepspaces=true,                 
    numbers=left,                    
    numbersep=5pt,                  
    showspaces=false,                
    showstringspaces=false,
    showtabs=false,                  
    tabsize=2
}
\newcommand\numberthis{\addtocounter{equation}{1}\tag{\theequation}}
\newcommand\scalemath[2]{\scalebox{#1}{\mbox{\ensuremath{\displaystyle #2}}}}

\usepackage[capitalize]{cleveref}
\crefname{section}{Sec.}{Secs.}
\Crefname{section}{Section}{Sections}
\Crefname{table}{Table}{Tables}
\crefname{table}{Tab.}{Tabs.}

\journal{Neurocomputing}

\begin{document}

\begin{frontmatter}

\title{Intra-class Patch Swap for Self-Distillation}

\author[label1]{Hongjun~Choi$^1$}

\author[label2]{Eun~Som~Jeon\footnote[1]{These authors contributed equally to this work.}}

\author[label3]{Ankita~Shukla}
\author[label4]{Pavan~Turaga}

\affiliation[label1]{organization={Lawrence Livermore National Laboratory},
             city={Livermore},
             postcode={94550},
             state={CA},
             country={USA}}
\affiliation[label2]{organization={Department of Computer Science and Engineering, Seoul National University of Science and Technology},
             city={Seoul},
             postcode={01811},
             country={South Korea}}
\affiliation[label3]{organization={Department of Computer Science and Engineering, University of Nevada},
             city={Reno},
             postcode={89557},
             state={NV},
             country={USA}}
\affiliation[label4]{organization={Geometric Media Lab, School of Arts, Media and Engineering and School of Electrical, Computer and Energy Engineering, Arizona State University},
             city={Tempe},
             postcode={85281},
             state={AZ},
             country={USA}}

\begin{abstract}
Knowledge distillation (KD) is a valuable technique for compressing large deep learning models into smaller, edge-suitable networks. However, conventional KD frameworks rely on pre-trained high-capacity teacher networks, which introduce significant challenges such as increased memory/storage requirements, additional training costs, and ambiguity in selecting an appropriate teacher for a given student model. Although a teacher-free distillation (self-distillation) has emerged as a promising alternative, many existing approaches still rely on architectural modifications or complex training procedures, which limit their generality and efficiency.

To address these limitations, we propose a novel framework based on teacher-free distillation that operates using a single student network without any auxiliary components, architectural modifications, or additional learnable parameters. Our approach is built on a simple yet highly effective augmentation, called intra-class patch swap augmentation. This augmentation simulates a teacher-student dynamic within a single model by generating pairs of intra-class samples with varying confidence levels, and then applying instance-to-instance distillation to align their predictive distributions. Our method is conceptually simple, model-agnostic, and easy to implement, requiring only a single augmentation function. Extensive experiments across image classification, semantic segmentation, and object detection show that our method consistently outperforms both existing self-distillation baselines and conventional teacher-based KD approaches. These results suggest that the success of self-distillation could hinge on the design of the augmentation itself. Our codes are available at https://github.com/hchoi71/Intra-class-Patch-Swap.

\end{abstract}


\end{frontmatter}



\section{Introduction}\label{sec1}
Deep neural networks have made impressive progress on a wide range of tasks, but they still struggle with overfitting and poor generalization due to the large number of parameters they have. To overcome these issues, knowledge distillation (KD) acts as a powerful regularization method by leveraging the complex pre-trained teacher's ability to convey learned logit and/or feature representations from the teacher network to the student network \cite{hinton2015distilling, zagoruyko2016paying, tung2019similarity, romero2014fitnets, tan2022improving}. This way, additional knowledge beyond the conventional learning from hard targets (i.e., one-hot labels) guides the student network to find better solutions and accelerate convergence.
\begin{figure}[htb!]
	\centering
	\includegraphics[width=0.7\linewidth]{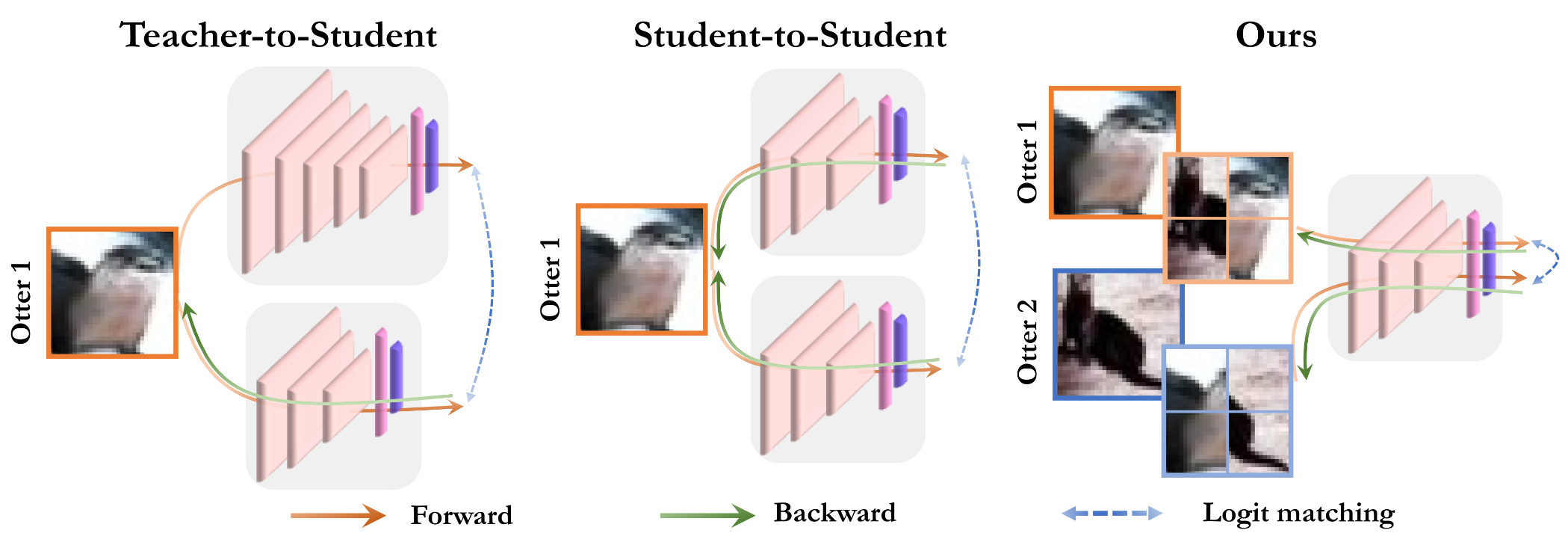}
	\caption{Three different distillation mechanisms. Teacher-to-Student and Student-to-Student require multiple networks, while self-distillation (ours) needs a single network to train it.}\label{diagram_mechanism} 
\end{figure}

Meanwhile, Zhang \textit{et al.} \cite{zhang2018deep} proposed online student learning, where the ensemble of students collaboratively learn from each other. The success of this collaborative learning strategy is related to increasing each student's posterior entropy when transferring knowledge from the Student-to-Student network. However, training a high-capacity teacher and multiple students (Teacher-to-Student and Student-to-Student in Figure \ref{diagram_mechanism}) is always cumbersome due to the need for high computational resources. Moreover, there is no universal standard for determining which teacher/student networks are the most suitable for distilling student networks, which often makes it challenging to select appropriate teacher networks. That is, there is an ambiguity in selecting the ``best-fit'' teacher for a given student.
Even poorly-trained teachers can improve the student's performance \cite{yuan2020revisiting}.

To address this issue, teacher-free distillation (self-distillation) approach has been introduced \cite{yuan2020revisiting, zhang2019your, moslemi2024survey}.
Self-distillation approach is to train a single model based on knowledge distillation, where the model behaviors as both the teacher and the student, and uses its own knowledge in training process. The goal is to match or even surpass the performance of conventional teacher-student KD methods without the need for an external teacher. However, several prior self-distillation methods require architectural modifications such as adding auxiliary classifiers which can be difficult to generalize across diverse backbone networks \cite{zhang2019your, moslemi2024survey}. In addition, some approaches introduce complex training procedures, including meta-learning or multi-stage optimization, which increase both training cost and memory footprint \cite{moslemi2024survey}.



\begin{figure*}[htb!]
	\centering
	\includegraphics[width=1.0\linewidth]{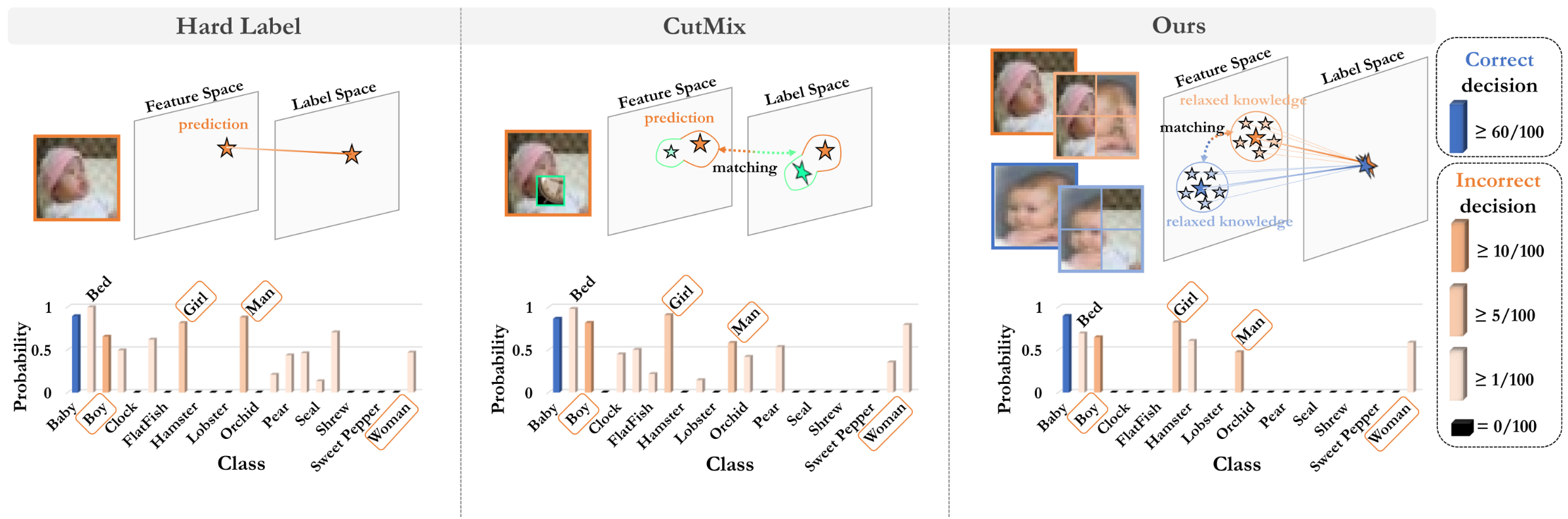}
	\caption{\textbf{Top: }Illustrations of design choices when learning the single network (ResNet-18) on CIFAR100. Instead of matching function directly (Hard Label, CutMix), our method tries to match the outputs (called relaxed knowledge) coming from two swapped inputs while we still use cross-entropy loss with the hard label. \textbf{Bottom: }Averaged top-1 probabilities of the target label from the test data. The blue and orange graphs indicate the averaged top-1 probability of correctly classified samples and misclassified samples, respectively. Compared to other methods, our method guarantees a high quality of predictive distribution, judging by the fact that even mispredictions still belong to the people class.}\label{Brief_oerview_of_proposed_method}
\end{figure*}

This study addresses a central question: How can we enable efficient and truly teacher-free distillation without relying on auxiliary components or architectural modifications, while maintaining applicability to diverse network types and achieving substantial performance gains?
To this end, intra-class samples are used, where one instance shares its knowledge with another instance in the same class. This sharing of instance-to-instance knowledge allows them to learn from each other as training progresses. However, this training approach may restrict performance improvement as it tends to prematurely converge toward local minima which are closely tailored to the training data \cite{chaudhari2019entropy, pereyra2017regularizing, keskar2016large}. To deal with this limitation, we further suggest an intra-class patch swap that gives and takes patches from input pairs from the same class. We present critical intuitions of why this is suited for the proposed self-distillation and what aspects of the designed augmentation differ from other popular data augmentations. 

\noindent\textbf{Generating event difficulty: }As a deep network focuses on the most discriminative part of the image, it is important to let the network see other relevant parts to broaden the understanding of the object \cite{kumar2017hide}. Motivated by this, we randomly exchange patches between two images in the same class. This could result in one image containing a strong teaching signal, such as the head of an object, while the other contains a relatively weak signal, such as the body. Then, the strong teaching signals generate highly confident knowledge about the true class, similar to that provided by the high-capacity teacher. In contrast, weaker knowledge reflects a lower confidence level. Matching these two different types (strong and weak) of knowledge results in a higher loss contribution compared to matching raw inputs. In other words, more weight is assigned to examples that show a substantial difference in confidence.

\ul{Throughout the training process, the random swap of patches within intra-class samples keeps generating challenges, thereby preventing the dominance of `easy-to-learn' samples in the classification}. Here, the `easy-to-learn' sample, figuratively speaking, refers to the image, mostly covered by the strong teaching signal. 

\noindent\textbf{Preserving class relationship: }
Mix-based augmentations such as MixUp \cite{zhang2017mixup} and CutMix \cite{yun2019cutmix} have attempted to enhance the generalizability of the networks by providing new training samples that blends two random images and labels proportionally.
However, mixing two random images might be detrimental to the class relationship, which does not consider intra-class variations or similarities.
The histogram of Figure \ref{Brief_oerview_of_proposed_method} provides compelling evidence, which depicts the averaged top-1 predictive distribution of the baby class across all baby test data. CutMix generates a new example by combining the target image (baby) with a non-target image (clock). Being trained with these images and labels might result in inferior predictive distribution. One can see that CutMix assigns probabilities of non-target categories with clock, flatfish, and pear (shown in orange), which are semantically unrelated to the target class. In contrast, \ul{our method produces the class probability distributed over semantically similar classes}, such as boy, girl, man, and woman. As we mentioned, behind KD's success lies the high quality of privileged information on a non-target class given by the powerful teacher \cite{choi2022knowledge}. An accurate teacher model guarantees a somewhat sophisticated class relationship, and we want to acquire such knowledge through distillation at the end. Since swapping the patches between positive samples does not change the intrinsic composition of the object, the class relationship is well-preserved. 

The contributions of this paper are as follows:
\begin{itemize}
\item We propose a simple yet highly effective data augmentation technique for self-distillation, an intra-class patch swap augmentation, which simulates a teacher-student dynamic within a single network by generating pairs of samples with varying prediction difficulty.
\item We introduce an instance-to-instance distillation framework that aligns the predictive distributions of patch-swapped samples. This approach captures the intrinsic composition of the target object and preserves semantic consistency, leading to more reliable learning representations. 

\item We extensively evaluate our method across multiple vision tasks such as image classification, semantic segmentation, and object detection, demonstrating its robustness, transferability, and superiority over both conventional KD and recent self-distillation baselines.
\end{itemize}



\section{Related Work}\label{sec:related} 
\subsection{Knowledge Distillation}

There are typically three groups of approaches for knowledge distillation during training: 

\noindent\textbf{Teacher-to-Student: }This process needs the supervision of the pre-trained teacher, and the student leverages the class relationship as captured by the teacher. They are also subdivided into two types based on the representation levels to be transferred, logit distillation from the final classifier \cite{hinton2015distilling} or the feature distillation from intermediate layers in networks \cite{park2019relational, tung2019similarity, zagoruyko2016paying}.

\noindent\textbf{Student-to-Student: }Instead of involving the large teacher during training, Zhang \textit{et al.} \cite{zhang2018deep} proposed mutual learning distillation, where it starts with a pool of untrained student models who learn from each other to solve the task together. Interestingly, this strategy achieved better performance than the conventional Teacher-to-Student method while growing in each other's knowledge. 

\noindent\textbf{Self-distillation: }Self-distillation has been proposed as an effective and efficient method to enhance knowledge distillation by generating self-knowledge. For instance, BYOT \cite{zhang2019your} adds classifiers in multiple latent blocks to diversify model parameters and features by leveraging their outputs in training the network. Additionally, Yuan \textit{et al}. conducted a theoretical analysis of the effectiveness of KD from the perspective of label smoothing regularization \cite{yuan2020revisiting} and proposed a teacher-free knowledge distillation (TF-KD) method that utilizes the label-smoothing effect. Another approach, OLS \cite{zhang2021delving}, updates the moving label distribution for each category during training to reflect the relationships between target and non-target categories. Moreover, data augmentation has been used for self-distillation to learn diverse views of samples. DDGSD \cite{xu2019data} attempts to transfer knowledge between two augmented outputs generated from a single sample (random mirror and cropping used as augmentation). DLB \cite{shen2022self} uses random crop and horizontal flip to create two sets of distorted mini-batches. It then employs the predictions of a network on the last mini-batch data in each epoch as a teacher to enforce consistency in predictions from the current mini-batch. CS-KD was also introduced, which matches the consistency between predictions of two batches of samples from the same category \cite{yun2020regularizing}. Specifically, CS-KD achieves this by detaching the gradient computation for one of the intra-class samples, minimizing the divergence between their outputs to mitigate the risk of model collapse. 

Unlike CS-KD, we perform full backpropagation through both intra-class samples. This preserves a richer learning signal, as the detached sample in CS-KD may not reliably provide the same high-quality supervision as a pre-trained teacher in traditional KD. Furthermore, existing self-distillation approaches have overlooked the critical role of designing effective data augmentations, often focusing primarily on the type of knowledge being transferred while neglecting the augmentation in the distillation itself. Despite growing interest in the interplay between data augmentations and the knowledge distillation (teacher-student frameworks), little attention has given to how data augmentation can be effectively utilized within self-distillation. In response, our work aims to design a simple yet effective data augmentation for self-distillation, demonstrating its significant impact.

\subsection{Data Augmentation} 
Here, we provide popular data augmentation techniques used in image classification, nothing that they are not intended as direct one-to-one comparisons. Instead, we found that by combining the strength of our method with other augmentations, additional performance gains can be achieved, as further discussed in Section \ref{analysis_aug}.

\noindent\textbf{Mix-based augmentations: }
MixUp \cite{zhang2017mixup} and CutMix \cite{yun2019cutmix} are known to be effective augmentation techniques that help to regularize network training by randomly mixing samples based on a randomly selected class. Cutmix, in particular, randomly cuts and pastes specific regions in a mini-batch to create mixing images and labels.
Recently, mix-based methods are adopted in knowledge distillation process to increase difficulty of understanding samples and to obtain better regularization \cite{yang2022mixskd, ijcai2021p319}. However, they require multiple networks or stages to leverage a pre-trained model and feature alignment modules additionally, which increase complexity and cost of resources for training process and need sensitive parameter settings.
Also, these augmentations mix class information and may not specifically target intra-class feature diversity, thus these does not inherently consider intra-class similarity or non-similarity.

Our patch swap technique differs from previous methods in several ways: 1) Our method always ensures that patches are swapped within intra-class examples, without discarding any part of the images, thereby preserving object integrity and encouraging the model to learn from varying representations within the same class rather than just random perturbations in different classes. 2) Randomly cropped regions can lead to noise labels since they may contain insignificant information, whereas the patch swap between positive samples reduces noise label concerns. 3) Ours does not require additional networks for feature alignment and not increase complexity in training. 


\noindent\textbf{Fine-grained augmentations: }  In the context of augmentation techniques applied during the distillation process, prior research \cite{wang2019distilling, wang2020fully} emphasizes the generation of fine-grained features guided by the teacher network. These studies operate on the premise that teacher networks possess superior capabilities for producing well-generalized feature responses. Then, the student model is encouraged to mimic the finely tuned features provided by the teacher network. This suggests that a well-crafted feature representation can significantly benefit the student model's performance in specific tasks. However, our approach diverges from this perspective. Rather than relying on superior knowledge from a teacher, we focus on devising simple yet effective augmentations that empower a single student model to acquire generalizable features, thereby enhancing its performance across tasks. 

Furthermore, using an attention network with swapping technique for fine-grained image classification was introduced to identify the discriminative part of an image \cite{zhang2021intra}, which is then swapped with the corresponding part from another image, and fed into the main classification network. While this swapping operation occurs within the same class, it's important to note that our method doesn't necessarily swap the most discriminative part only. Within our proposed framework, the key objective of patch swapping is to consistently generate varying difficulty levels for two samples within the intra-class instances. Therefore, it subsequently achieves a balance between `hard-to-learn' and `easy-to-learn' samples through the distillation process. As a result, we do not require an additional attention module to detect the discriminative part of the object, which helps to avoid an increase in computational resources.


\section{Proposed Method}\label{sec:method}
\subsection{Preliminaries}
We define the input and its label as $\mathcal{D}=\{({x}_{i}, y_{i})\}^{N}_{i=1}$ where $y_{i}$ is the hard label of the input ${x}_{i}$, and $N$ is the number of training samples. To leverage the intra-class pairs in our method, we denote the first mini-batch input as $\mathcal{B}_{a}=\{({x}_{ai}, y_{ai})\}^{n}_{i=1}$ and $\mathcal{B}_{b}=\{({x}_{bi}, y_{bi})\}^{n}_{i=1}$ for the second one, where $n$ is the batch size. Note, the intra-sample pairs guarantees $y_{a}=y_{b}$. Deep neural networks produce the logits $f_{i}$ that can be converted into the probability $p_{i}$ using the softmax function as follows: $p_{k}=\frac{\exp(f_{k}/\mathcal{T})}{\sum_{j}\exp(f_{j}/\mathcal{T})}$, the predicted value of class $k\in\mathcal{C}$ where $\mathcal{C}$ is the total number of class. Here, the temperature $\mathcal{T}$ is to modulate the smoothness in the probability distribution.  

\subsection{Intra-class Patch Swap}\label{3.2}

\begin{figure*}[htb!]
	\centering
	\includegraphics[width=1.0\linewidth]{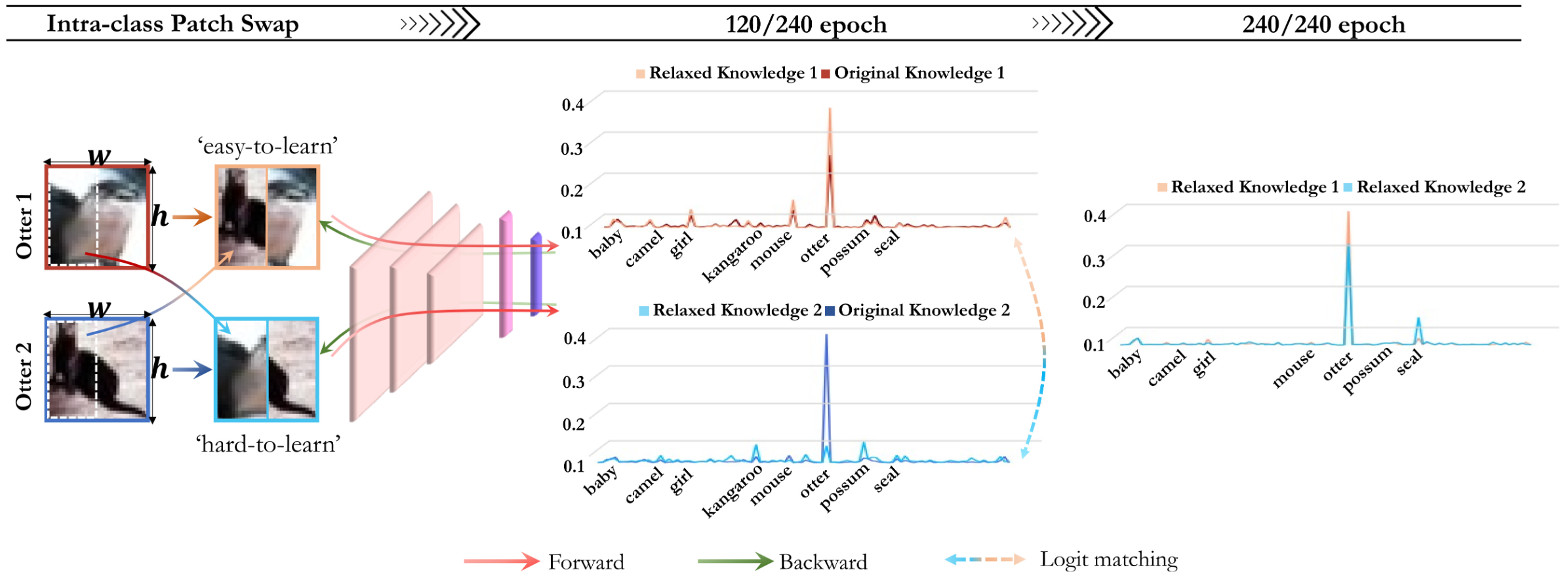}
	\caption{Overall framework of the proposed method. The process of generating new training inputs by exchanging patches between positive sample pairs creates a strong teaching signal, leading to high prediction confidence in the target class (Relaxed Knowledge 1) and low prediction confidence in the same target for the image that lost the strong signal (Relaxed Knowledge 2). By matching these outputs, the network learns all relevant parts of the object, as demonstrated by a well-matched graph in the final figure.}\label{overall_framework}
\end{figure*}

In this section, we present the process of the intra-class patch swap and its effect with two following aspects:

\noindent\textbf{1) Event difficulty generation: }
We exchange patches between two images in the same class randomly, which possibly results in strong teaching signals. 
For example, as describe in Figure \ref{overall_framework}, strong signals (e.g., the head of the otter) being present only in one image, while another image contains only weak signals such as the body or tail of the otter. This can lead to an increase in the difference in event difficulty between the outputs. In halfway during the training process, the output distribution produced by the first input becomes more sharply peaked on the target class as the strong supervisory signals feed into the network (see the change in confidence on the otter class from Original Knowledge 1 to Relaxed Knowledge 1). In contrast, the output distribution for the second input becomes low confidence on the target due to the loss of strong signals (from Original Knowledge 2 to Relaxed Knowledge 2). When matching between Relaxed Knowledge 1 and 2, the loss contribution to these examples grows, which implies that the network focuses on the examples showing high differences in event difficulty, and it simultaneously tries to learn the relevant parts of the object rather than focusing on the most strong supervisory parts.

\noindent\textbf{2) Preserving class relationship: }Another important role of the random swap is that it does not hurt the class relationship among different categories. For intuitive understanding, we denote the newly generated first input as the `easy-to-learn' sample in Figure \ref{overall_framework} because the distribution sharply peaks on the target than the distribution from the raw input. Similarly, we denote the generated second input as the `hard-to-learn' example that consists of non-discriminative regions. Even if the second input loses strong supervisory signals, it plays a critical role in delivering useful information in that it emphasizes the probability distributions for semantically similar classes (kangaroo, mouse, possum, and seal) other than the target class. This is represented by the light blue graph in the middle column of the figure. It is worth noting that this aspect results in an improved quality of predictive distribution grouping for semantically similar classes. We provide a more comprehensive analysis of the predictive distribution, both quantitatively (see Section \ref{analysis_aug}) and qualitatively (see Section \ref{qual_pred}) \\

\noindent\textbf{Interplay `easy-to-learn' and `hard-to-learn': }In the distillation process, exposing the student network to diverse knowledge synthesized by the teacher network has been emphasized \cite{gou2021knowledge, wen2021preparing, jeon2023leveraging}. Building upon this insight, the proposed strategy involves the \textit{\textbf{random swapping}} of different patches between pairs of intra-class inputs. This strategy plays a central role in guiding the network to concentrate on various relevant aspects of the object, extending its focus beyond the most discriminative parts during training. As shown in the last graph of Figure \ref{overall_framework}, when the training reaches the end, the network is grown to understand all of the relevant parts of the object by matching Relaxed Knowledge 1 and 2. Two well-matched graphs represent this phenomenon. 

The interplay between `easy-to-learn' and `hard-to-learn' is highly effective in the self-distillation framework. By swapping random patches in every iteration of training, different event levels are constantly created, resulting in a loss calculated by the swapped image that maintains a relatively larger value than the loss calculated with the raw image. This indicates that the model is penalized heavily for hard samples, preventing the easy samples from overwhelming the classification during training. Furthermore, we confirm that the proposed swap alone (i.e., using only the first swapped inputs to train the model with cross-entropy loss) does not contribute to network performance improvement (as observed in the ``Hard Label+Swap'' in Section \ref{analysis_aug}). Therefore, it is apparent that patch swapping exhibits its effectiveness when applied to intra-class sample pairs. 

It is also important to note that while Figure \ref{overall_framework} designates two intra-class samples as either easy or hard for clarity, \ul{the randomness employed in swapping does not always ensure that one sample is a `hard-to-learn' example while the other is an `easy-to-learn' one}. To understand the influence of generating event difficulty, we present the confidence gap between two predictive probabilities on the target class throughout the training process in Section \ref{effect_patch}.

\noindent\textbf{Swap intra-class patches: }Given intra-class sample pair, ${x}_{i}$ and ${x}_{j}$ of size $w\times h$, we first divide the input into a grid with a fixed patch size of $m$. The image eventually has $w/m\times h/m$ patches in total. Importantly, the set of patches in an image is newly chosen based on the randomness at every iteration, and they exchange to the same location in another image. We denote the swapped inputs as $\hat{x}_{i}$ and $\hat{x}_{j}$. Note, $p_{r}$ determines whether we use the swap augmentation or not in a given pair from the batch. In the case of $p_{r}=0$, no swapping happens. Though we set $p_{r}$ to $0.5$ in all experiments, we will investigate the effect of different values $p_{r}$ in Section \ref{effect_hyperparameter}.

\subsection{Self Distillation}
Here, we describe the objective function for training our network. $L_{c}$ denotes the cross-entropy loss with one-hot encoding labels. This is still important to optimize the network in order to perform well in classifying the image.
\begin{align*}
\scalemath{0.90}{\mathcal{L}_{C1}(f^{S},y)=-\sum_{i=1}^{n}y_{ai}\log(\sigma(f^{S}(\hat{x}_{ai}))}, \\
\scalemath{0.90}{\mathcal{L}_{C2}(f^{S},y)=-\sum_{i=1}^{n}y_{bi}\log(\sigma(f^{S}(\hat{x}_{bi}))}, \numberthis \label{eq:ce}
\end{align*}
where $y_{ai}$ and $y_{bi}$ are same one-hot labels. $1$ and $2$ attached to $\mathcal{L}_{C}$ denote the respective loss from two swapped inputs. And $\sigma$ indicates the softmax function. To reach consistent predictive distributions from two inputs, we use $KL$ divergence between them as followings: 
\begin{align*}
\scalemath{0.8}{\mathcal{L}_{KD1}(f^{S},f^{S})=\frac{1}{n}\sum_{i=1}^{n}\mathcal{T}^{2}KL(\sigma(\frac{f^{S}(\hat{x}_{ai})}{\mathcal{T}}), \sigma(\frac{f^{S}(\hat{x}_{bi})}{\mathcal{T}}))}, \\
\scalemath{0.8}{\mathcal{L}_{KD2}(f^{S},f^{S})=\frac{1}{n}\sum_{i=1}^{n}\mathcal{T}^{2}KL(\sigma(\frac{f^{S}(\hat{x}_{bi})}{\mathcal{T}}), \sigma(\frac{f^{S}(\hat{x}_{ai})}{\mathcal{T}}))}, \numberthis \label{eq:selfkd_eq1}
\end{align*}
\noindent where $KL(A,B)=-\sum_{x\in\mathcal{X}}A\log(\frac{B}{A})$. The higher temperature renders the outputs softer so that the higher class relationship among categories injects into the output distribution. We empirically found that $\mathcal{T}=4$ leads to the best performance. Finally, the above two equations can be incorporated to update the network parameters via stochastic gradient optimization. 
\begin{align*}
\mathcal{L}=\frac{1}{2}\gamma(\mathcal{L}_{C1}+\mathcal{L}_{C2})+\frac{1}{2}\alpha(\mathcal{L}_{KD1}+\mathcal{L}_{KD2}), \numberthis \label{eq:selfkd_eq}
\end{align*}
where $\gamma$ and $\alpha$ are hyper-parameters adjusting contributions in each component. We assign identical values to ensure equivalent importance between the cross-entropy loss and the $KL$ loss. We present the pseudo-code of mini-batch training in Algorithm \ref{Alg:mini-batch_training_swap}. 

\begin{algorithm}[htb!]
\caption{{Mini-batch training}\label{Alg:mini-batch_training_swap}}
\textbf{Require: } $B_{a}$ and $B_{b}$ = Mini batch from training set, $\mathcal{D}$ \\
\textbf{Require: } $f_{\theta}^{S}(x)$ = Single network with parameter $\theta$ 
\begin{algorithmic}[1]
\For{$x_{a}\in B_{a}$, $x_{b}\in B_{b}$}
\State $\hat{x}_{a}$, $\hat{x}_{b}$ = \textbf{Intra\_Patch\_Swap}($x_{a}$, $x_{b}$, $m$, $p_{r}$)
\State ${f}^{S}_{a}$ $\leftarrow$ $f^{S}(\hat{x}_{a})$
\State ${f}^{S}_{b}$ $\leftarrow$ $f^{S}(\hat{x}_{b})$
\State Compute the joint loss based on Eq \ref{eq:selfkd_eq}.
\State Update $\theta$ using SGD optimizers
\EndFor
\State return $\theta$
\end{algorithmic}
\end{algorithm}

We also provide the PyTorch-style full code of Intra-class Patch Swap. We incorporate this function into the mini-batch training process, as outlined in the algorithm. In line 12, the set of patches in an image is selected randomly, and this process ensures that at least one patch in the first input is swapped with the patch in the second input, but the number of selected patches should not exceed $w/m\times h/m-1$. As seen in the code, our proposed patch swap for self-distillation is straightforward and easy to implement.

\begin{lstlisting}[language=Octave, basicstyle=\ttfamily\footnotesize, numbers=left, xleftmargin=2em]
import torch as T
import torch.nn.functional as F
'''
Input: 1) Sample mini batch B_a and B_b from training and label set, D
       2) S: w // int(np.sqrt(t_p)) where w is the width of a image and t_p is the total patches
'''
# Divide each mini-batch into non-overlapping patches
x_a = F.unfold(B_a, kernel_size=S, stride=S, padding=0); p_xa=[]
x_b = F.unfold(B_b, kernel_size=S, stride=S, padding=0); p_xb=[]
for (xai, xbi) in zip(x_a, x_b):
    if np.random.rand(1)<p_r:
            idx = T.randperm(t_p)[:T.randint(1,t_p,(1,))[0]]
            xai[:,idx], xbi[:,idx] = xbi[:,idx], xai[:,idx]
            p_xa.append(xai[None,...])
            p_xb.append(xbi[None,...]) 
    else: "No Swap"
# Back to the origianl form of inputs
xai_hat = F.fold(p_xa, B_a.shape[-2:], kernel_size=S, stride=S, padding=0)
xbi_hat = F.fold(p_xb, B_b.shape[-2:], kernel_size=S, stride=S, padding=0)

\end{lstlisting}

\subsection{Analyzing the Effect of Patch Swap} \label{effect_patch}
In this section, we explore possible explanations for performance improvements in self-distillation with the patch swap.

\begin{figure}[htb!]
	\centering
	\includegraphics[width=0.6\linewidth]{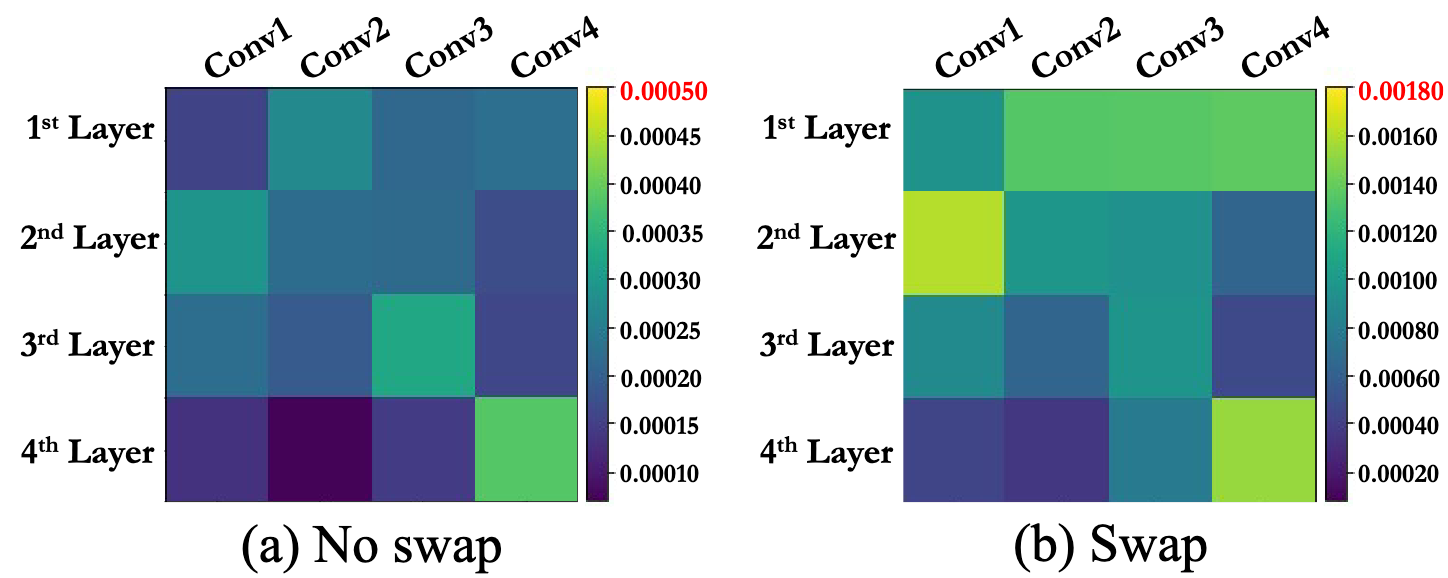}
	\caption{Magnitude of gradients in each layer from ResNet18. Patch swap, as used in our method, can indirectly alleviate the vanishing gradient problem that may arise during training.}\label{van_grad}
\end{figure}

\noindent\textbf{Vanishing gradient: } Vanishing gradients are a common problem encountered in training, which occurs when the gradient of the loss function with respect to the parameters becomes exceptionally small. As a result, the optimizer has difficulty effectively updating the model's parameters. To assess the effect of patch swapping on this issue, we calculated the average absolute value of gradients in each convolutional layer of ResNet18 on CIFAR100, as shown in Figure \ref{van_grad}. Our analysis indicates that the magnitude of gradients without patch swap is very small, thus the network is unable to update the weights in these layers effectively. This can result in these layers being underutilized and not contributing to the performance of the network. In contrast, when employing patch swap, our observations reveal a notable increase in gradient magnitudes. This enhancement allows for more effective weight updates within these layers, ensuring their active participation and significant contribution to the overall network performance.

\begin{figure}[htb!]
	\centering
	\includegraphics[width=0.5\linewidth]{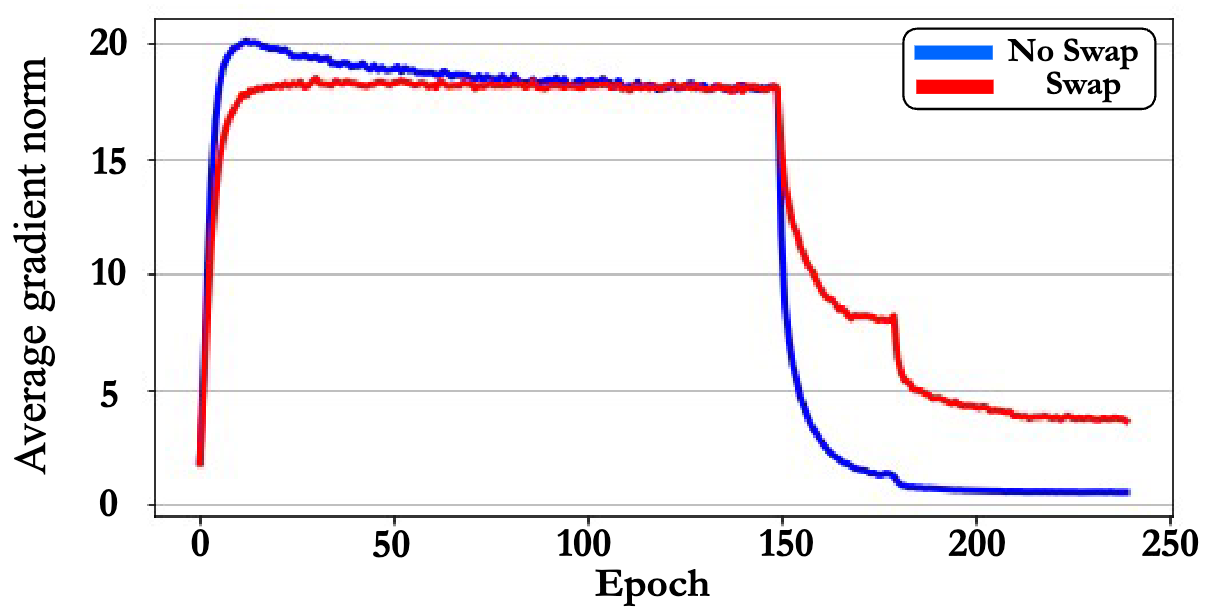}
	\caption{The averaged L1 norm of the gradient. Patch swap keeps relatively high values after 150 epochs, which suggests that the model continues to learn throughout the training process.}\label{avg_grad}
\end{figure}

\noindent\textbf{Monitoring the trend in training: } We claim that the proposed swap constantly generates event difficulty during training, thus our method hinders the model from getting overwhelmed with ‘easy-to-learn’ examples. To support our claim, we analyze the gradient of the $\mathcal{L}_{KD1}$ with ResNet18 \cite{he2016deep} on CIFAR100 \cite{krizhevsky2009learning} as follows: $\partial\mathcal{L}_{KD1}/\partial f_{k}=p(\hat{x}_{bk})-p(\hat{x}_{ak})$, where, $p(\hat{x}_{ak})$ and $p(\hat{x}_{bk})$ are the model's probabilities confidence on the target $k$ for the first and the second swapped input, respectively. Then, we average the L1 norm of the gradient across training samples. Figure \ref{avg_grad} shows that the effect of gradient rescaling is related to event difficulty. The swap has a stronger effect on the gradient than the raw input. This is a potential indication that the model without patch swap has prematurely converged, resulting in suboptimal performance.

\noindent\textbf{Training and testing accuracy gap: }In the study on the effects of batch size, Keskar \textit{et al.} demonstrated empirically that using a large batch size makes the gradient estimate less noisy but biased towards the most frequent direction of the gradients. As a result, the optimization algorithm may get stuck in sharp minima with a small region of attraction \cite{keskar2016large}. Conversely, using a small batch size introduces more noise into the gradient estimate, which can help the algorithm escape from sharp minima and find flat minima that generalize better, as evidenced by the performance gap between training and testing accuracy. 

\begin{table}[htb!]
\renewcommand{\tabcolsep}{1.2mm} 
\caption{Performance of the proposed model without/with the intra-class swap.}
\label{test_accuracy_gap}
\centering 
\begin{center}
\begin{tabular}{@{} c *{4}{S} @{}} 
\toprule
 & \multicolumn{2}{c}{\makecell{Training Accuracy}} &
\multicolumn{2}{c}{\makecell{Testing Accuracy}} \\ 
\cmidrule(lr){2-3}\cmidrule(l){4-5}
Model & {No Swap} & {Swap} & {No Swap} & {Swap} \\
\midrule
ResNet18\cite{he2016deep} & 99.97\% & 98.48\% & 79.18\% & 80.33\% \\
ResNet34\cite{he2016deep} & 99.96\% & 95.31\% & 79.22\% & 81.18\% \\
VGG13\cite{simonyan2014very} & 99.96\% & 96.98\% & 75.65\% & 77.28\% \\
VGG16\cite{simonyan2014very} & 99.93\% & 96.70\% & 74.86\% & 77.39\% \\
ShuffleNetV1\cite{ma2018shufflenet} & 99.54\% & 95.32\% & 71.29\% & 74.15\% \\
ShuffleNetV2\cite{ma2018shufflenet} & 99.59\% & 95.27\% & 72.75\% & 75.40\% \\
\bottomrule
\end{tabular}
\end{center}
\end{table}

Our framework requires two intra-class examples as input, doubling the batch size. This has a similar effect to the observations made in previous studies \cite{keskar2016large}. To investigate the behavior of our model with and without patch swap, we adopt an analytical tool adopted from \cite{keskar2016large}. Consistent with Keskar's observations, we also found limited performance gain when training our model when doubled batch feeds into the network without patch swap. As shown in Table \ref{test_accuracy_gap}, while both models achieve high training accuracy, the model trained with patch swap shows significant improvements on the test set and yields a smaller gap between training and testing accuracy compared to the model without patch swap. Our proposed patch swap method indirectly alleviates the issue of finding sharp minima that may arise during training.

\section{Experiments}\label{sec:experiment}
We validate the effectiveness of our proposed method on image classification, semantic segmentation, and object detection, and further evaluate its safety using various metrics.   

\noindent\textbf{Reproducibility: } The code capable of reproducing all experiment results will be available on personal GitHub page.

\subsection{Compared Methods} Our experimental comparisons primarily focus on existing self-distillation methods, as our approach builds upon the concept of knowledge discrepancy rather than directly enhancing data augmentation. This choice aligns with the core principles of our approach: 1) Core mechanism: Our approach is motivated by a knowledge discrepancy between a teacher network (higher confidence) and a student network (lower confidence). Intra-class swap naturally creates this discrepancy by exchanging patches within samples of the same class, eliminating the need to explicitly train the teacher. 2) Augmentation context: Intra-class swap is not a standalone, achieving SOTA augmentation results. When training a single network without Eq. \ref{eq:selfkd_eq} (typical way of training with the cross-entropy loss), intra-class swap offers negligible improvements (approximately 0.1\% accuracy gain). 

To provide broader context, we also include comparisons with popular data augmentation techniques (MixUp, CutMix, CutOut) and investigate their compatibility with intra-class swap.  We categorize baseline methods into the following groups:

\noindent\textbf{Label regularization: }Label Smoothing (LS) \cite{szegedy2016rethinking} and Disturb Label \cite{xie2016disturblabel} refine labels by smoothing or adding noise, training the network using these labels instead of hard labels. 

\noindent\textbf{Self-knowledge distillations: } Methods in this category diversify and leverage network predictions from a single network. This includes:

\begin{itemize}
  \item BYOT \cite{zhang2019your}: Adds multiple classifiers to ResNet blocks for diversified predictions, minimizing differences during training.
  \item TF-KD \cite{yuan2020revisiting}: Incorporates LS and uses a manually designed virtual teacher with 100\% accuracy.
  \item CS-KD \cite{yun2020regularizing}: Enforces consistency between predictions of two samples in the same class, while detaching the gradient computation of one sample (similar to an offline teacher). 
  \item DDGSD \cite{xu2019data}: Leverages outputs from differently augmented versions of a sample (e.g., random crops and flips) to produce diverse predictions, enforcing consistency between them.
  \item DLB \cite{shen2022self}: Enforces consistency between augmented outputs produced by the last mini-batch data and the current mini-batch data.
  \item OLS \cite{zhang2021delving}: Creates smooth target distributions using a moving average scheme, encouraging smoother model predictions.
  \item FER \cite{deng2021fer}: Limits the network's behavior change on correctly classified samples, aiming to maintain learned classification boundaries. 
\end{itemize}

\noindent\textbf{Teacher-student distillation: } Methods utilizing knowledge transfer from a powerful teacher to a smaller student network. Examples include:

\begin{itemize}
  \item AT \cite{zagoruyko2016paying}: Leverages attention maps generated by a powerful teacher network to boost performance in a smaller student network.
  \item SP \cite{tung2019similarity}: Transmits knowledge from teacher to student network by enforcing similarity between their output distributions.
  \item RKD \cite{park2019relational}: Focuses on transferring relationships between data points from a teacher to a student network.
  \item CRD \cite{tian2019contrastive}: Improves student networks by having them learn representations that pull similar samples together and push dissimilar ones apart in a way of teacher learning its representations.

\end{itemize}

\subsection{Image Classification}
\noindent\textbf{Implementation details: } We use multiple datasets and network architectures for our experiments, including CIFAR100 \cite{krizhevsky2009learning}, ImageNet \cite{deng2009imagenet}, CUB-200-2011 \cite{wah2011caltech}, and Stanford Dogs \cite{khosla2011novel}. All methods are trained for 240 epochs on CIFAR100 with a learning rate of 0.05, decayed by a factor of 0.1 every 30 epochs after 150 epochs. For ShuffleNet V1 and V2, the initial learning rate is set to 0.01. The balancing parameters $\gamma$ and $\alpha$ are set to 1.0 with $4\times4$ patches and $p_{r}=0.5$, and the batch size is set to 64. We also tried setting $\gamma=0.5$ and $\alpha=0.5$, which worked better for higher-capacity networks like ResNet34, ResNet50, ResNet101, and VGG16, as smaller weighing on the loss leads to slightly better convergence in the early training stage. For ImageNet in Table \ref{test_accuracy_table_on_imagenet}, we train the ResNet50 model for 100 epochs with an initial learning rate of 0.1, decayed by a factor of 0.1 at 30, 60, and 90 epochs, and a batch size of 256. The balancing parameters $\gamma$ and $\alpha$ are both set to 1.0 with $p_{r}=0.5$. We also train MobileNetV2 \cite{sandler2018mobilenetv2} for 60 epochs to explore the applicability of the proposed method to diverse networks. The proposed method achieves improvements of 0.78\% in top-1 accuracy and 1.11\% in top-5 accuracy over the baseline. While these improvements are smaller than the gains observed for the heavier network (ResNet50), the compact model still benefits from enhanced performance through our method.
Furthermore, we test with an up-to-date transformer architecture, MViTv2 \cite{li2022mvitv2}, to show that the proposed framework can be applied to networks other than convolutional neural networks. To highlight the effect of the proposed patch swap, we exclude additional augmentations such as MixUp and CutMix, while maintaining alignment with CNN-based network settings: batch size of 192 and 200 epochs. With these results, we observe the generalizability of our method, which can be applied to various architectures and types of networks.

For all baseline methods, we used the same hyperparameter settings as those specified in their papers and ran them using their author-provided code. 

\begin{table}[htb!]
\renewcommand{\tabcolsep}{5.0mm} 
\centering
\caption{Top-1 and Top-5 classification accuracy (\%) on ImageNet. $\dagger$ denotes the results from \cite{zhang2021delving} and $\ddagger$ from \cite{liang2022efficient}. The best results are highlighted in \textbf{Bold}.}
\label{test_accuracy_table_on_imagenet}
\begin{center}
\begin{tabular}{lr}
\toprule
Model + Method & Top-1 / Top-5  \\
\midrule
ResNet50 + Hard Label & 76.30 / 93.05  \\
ResNet50 + LS\cite{szegedy2016rethinking} & 76.67 / -$^{\ddagger}$ \\
ResNet50 + CutOut\cite{devries2017improved} & 77.07 / 93.34$^{\dagger}$ \\
ResNet50 + Disturb Label\cite{xie2016disturblabel} & 76.41 / 93.10$^{\dagger}$ \\
ResNet50 + BYOT\cite{zhang2019your} & 76.96 / 93.49$^{\dagger}$ \\
ResNet50 + TF-KD\cite{yuan2020revisiting} & 76.56 / -$^{\ddagger}$ \\
ResNet50 + CS-KD\cite{yun2020regularizing} & 76.78 / -$^{\ddagger}$ \\
ResNet50 + Zipf’s LS\cite{liang2022efficient} & 77.25 /-$^{\ddagger}$ \\ 
\midrule
ResNet152 (Teacher)& \multirow{2}{*}{77.49 /-} \\
ResNet50 + KD\cite{hinton2015distilling} &  \\ 
\midrule
\multirow{2}{*}{ResNet50 + Ours (2$\times$2)} & \textbf{77.85} \, / \, \textbf{93.57} \, \\
& (\textcolor{ForestGreen}{1.55$\uparrow$}) / (\textcolor{ForestGreen}{0.52$\uparrow$}) \! \\

\multirow{2}{*}{ResNet50 + Ours (4$\times$4)} & \textbf{77.59} \, / \, \textbf{93.56} \, \\
& (\textcolor{ForestGreen}{1.29$\uparrow$})  /  (\textcolor{ForestGreen}{0.51$\uparrow$}) \! \\

\midrule
MobileNetV2 \cite{sandler2018mobilenetv2} & 60.05 \, / \, 83.20 \,   \\
\multirow{2}{*}{MobileNetV2 + Ours (2$\times$2)} & \textbf{60.83} \, / \, \textbf{84.31} \, \\
& (\textcolor{ForestGreen}{0.78$\uparrow$}) / (\textcolor{ForestGreen}{1.11$\uparrow$}) \! \\

\midrule
MViTv2 \cite{li2022mvitv2} & 77.71 \, / - \\
\multirow{2}{*}{MViTv2 + Ours (2$\times$2)} & \textbf{80.99} \, / -  \\
& (\textcolor{ForestGreen}{3.28$\uparrow$})   \! / -\\
\bottomrule
\end{tabular}
\end{center}
\end{table}

\begin{table*}[htb!]
\centering
\renewcommand{\tabcolsep}{1.3mm} 
\caption{Accuracy (\%) of the networks trained with several baseline methods on CIFAR100. The best results are indicated in \textbf{bold}. The reported top-1 accuracy is the average of three runs. The performance gap with Hard Label is shown in green.}
\label{test_accuracy_table_on_cifar100}
\begin{center}
\scalebox{0.95}{\begin{tabular}{l|cccccccc}
\toprule
Method & ResNet18 & ResNet34 & ResNet50 & ResNet101 & VGG13 & VGG16 & ShuffleNetV1 & ShuffleNetV2 \\
\midrule
Hard Label & 77.92\scriptsize$\pm$0.10 & 79.05\scriptsize$\pm$0.09 & 79.01\scriptsize$\pm$0.24 & 79.29\scriptsize$\pm$0.21 & 74.62\scriptsize$\pm$0.23 & 74.40\scriptsize$\pm$0.03 & 71.26\scriptsize$\pm$0.11 & 72.62\scriptsize$\pm$0.18 \\
LS\cite{szegedy2016rethinking} & 78.78\scriptsize$\pm$0.15 & 79.14\scriptsize$\pm$0.17 & 79.09\scriptsize$\pm$0.05 & 79.30\scriptsize$\pm$0.32 & 75.68\scriptsize$\pm$0.21 & 74.66\scriptsize$\pm$0.25 & 71.42\scriptsize$\pm$0.29 & 73.29\scriptsize$\pm$0.07 \\
Disturb Label\cite{xie2016disturblabel} & 78.25\scriptsize$\pm$0.25 & 78.51\scriptsize$\pm$0.43 & 78.53\scriptsize$\pm$0.59 & 79.47\scriptsize$\pm$0.66 & 75.10\scriptsize$\pm$0.11 & 74.37\scriptsize$\pm$0.15 & 71.89\scriptsize$\pm$0.08 & 73.05\scriptsize$\pm$0.07 \\
DDGSD\cite{xu2019data} & 79.32\scriptsize$\pm$0.06 & 79.46\scriptsize$\pm$0.38 & 80.62\scriptsize$\pm$0.47 & 80.98\scriptsize$\pm$0.10 & 76.47\scriptsize$\pm$0.09 & 75.53\scriptsize$\pm$0.28 & 71.97\scriptsize$\pm$0.21 & 73.91\scriptsize$\pm$0.16 \\
TF-KD\cite{yuan2020revisiting} & 79.05\scriptsize$\pm$0.29 & 78.88\scriptsize$\pm$0.07 & 79.15\scriptsize$\pm$0.95 & 79.84\scriptsize$\pm$0.12 & 76.13\scriptsize$\pm$0.04 & 74.99\scriptsize$\pm$0.14 & 71.93\scriptsize$\pm$0.20 & 73.11\scriptsize$\pm$0.23\\
CS-KD\cite{yun2020regularizing} & 78.85\scriptsize$\pm$0.15 & 79.25\scriptsize$\pm$0.09 & 78.99\scriptsize$\pm$0.31 & 79.28\scriptsize$\pm$0.36 & 76.10\scriptsize$\pm$0.26 & 74.92\scriptsize$\pm$0.33 & 72.26\scriptsize$\pm$0.32 & 73.26\scriptsize$\pm$0.16 \\
OLS\cite{zhang2021delving} & 78.95\scriptsize$\pm$0.10 & 79.34\scriptsize$\pm$0.28 & 80.07\scriptsize$\pm$0.34 & 80.43\scriptsize$\pm$0.15 & 75.69\scriptsize$\pm$0.08 & 74.54\scriptsize$\pm$0.13 & 72.34\scriptsize$\pm$0.13 & 73.44\scriptsize$\pm$0.30 \\
FER\cite{deng2021fer} & 79.99\scriptsize$\pm$0.18 & 80.00\scriptsize$\pm$0.23 & 80.35\scriptsize$\pm$0.21 & 79.99\scriptsize$\pm$0.30 & - & - & 73.09\scriptsize$\pm$0.13 & 74.52\scriptsize$\pm$0.18\\
DLB\cite{shen2022self} & 80.25\scriptsize$\pm$0.15 & 80.71\scriptsize$\pm$0.12 & 81.35\scriptsize$\pm$0.15 & 81.83\scriptsize$\pm$0.06 & 77.08\scriptsize$\pm$0.38 & 76.58\scriptsize$\pm$0.19 & 73.80\scriptsize$\pm$0.31 & 74.30\scriptsize$\pm$0.25 \\
\midrule
Ours (4$\times$4) & \makecell{\textbf{80.53}\scriptsize$\pm$\textbf{0.16} \\ (\textcolor{ForestGreen}{2.61$\uparrow$})} & \makecell{\textbf{81.61}\scriptsize$\pm$\textbf{0.24} \\ (\textcolor{ForestGreen}{2.56$\uparrow$})} & \makecell{\textbf{81.97}\scriptsize$\pm$\textbf{0.22} \\ (\textcolor{ForestGreen}{2.96$\uparrow$})} & \makecell{\textbf{82.71}\scriptsize$\pm$\textbf{0.12} \\ (\textcolor{ForestGreen}{3.42$\uparrow$})} & \makecell{\textbf{77.90}\scriptsize$\pm$\textbf{0.02} \\ (\textcolor{ForestGreen}{3.28$\uparrow$})} & \makecell{\textbf{77.66}\scriptsize$\pm$\textbf{0.11} \\ (\textcolor{ForestGreen}{3.26$\uparrow$})} &  \makecell{\textbf{74.32}\scriptsize$\pm$\textbf{0.29} \\ (\textcolor{ForestGreen}{3.06$\uparrow$})} & \makecell{\textbf{75.75}\scriptsize$\pm$\textbf{0.22} \\ (\textcolor{ForestGreen}{3.13$\uparrow$})}  \\
\bottomrule
\end{tabular}}
\end{center}
\end{table*}

\noindent\textbf{Comparison with `Self-Distillation' methods: } 
Table \ref{test_accuracy_table_on_imagenet} shows that we achieve the best top-1 and top-5 performance among others with 1.55\% improvements for top-1 on the ImageNet. Interestingly, our method also shows better performance than the conventional KD when using the teacher model (ResNet152). We observe that $2\times2$ patches slightly perform better than $4\times4$ ones. Also, the result clearly demonstrates that our proposed self-distillation method with MViTv2 surpasses the baseline (referred to as Hard Label, trained without distillation) by a substantial margin of 3.28\%. 
In Table \ref{test_accuracy_table_on_cifar100}, experimental results with the top-1 accuracy of various self-distillation methods on CIFAR100 are shown. As seen in this table, our proposed method significantly outperforms all baselines and shows consistent robustness against the network choice.

\noindent\textbf{Comparison with `Teacher-to-Student' methods: }The conventional knowledge distillation methods require a pre-trained teacher model to take an advantage of the power for their accurate knowledge to train a student model. This way speeds up the convergence rate and enhances the generalization ability of the network \cite{hinton2015distilling}. However, our study shows that the proposed method, running without additional networks or a teacher model, can produce strong enough knowledge to guide itself. We compare the performance of the proposed method with popular distillation methods under three different teacher-student combinations. As seen in Table \ref{test_accuracy_table_on_kd}, our self-distillation achieves better performance than other `Teacher-to-Student' baselines.

\begin{table}[htb!]
\centering
\renewcommand{\tabcolsep}{1.6mm} 
\caption{Accuracy (\%) on CIFAR100 dataset. Note, our method does not need the teacher network. The accuracy is averaged over three runs. The best results are indicated in \textbf{bold}.}
\label{test_accuracy_table_on_kd}
\begin{center}
\begin{tabular}{l|ccc}
\toprule
Teacher & ResNet50 & ResNet50 & ResNet101  \\
Student & ResNet18 & ResNet34 & VGG16  \\
\midrule
Hard Label & 77.92\scriptsize$\pm$0.10 & 79.05\scriptsize$\pm$0.09 & 74.40\scriptsize$\pm$0.03   \\
KD \cite{hinton2015distilling} & 79.64\scriptsize$\pm$0.06 & 79.99\scriptsize$\pm$0.25 & 76.36\scriptsize$\pm$0.07  \\
AT \cite{zagoruyko2016paying} & 78.30\scriptsize$\pm$0.03 & 79.25\scriptsize$\pm$0.06 & 74.87\scriptsize$\pm$0.21  \\
SP \cite{tung2019similarity} & 79.92\scriptsize$\pm$0.15 & 80.18\scriptsize$\pm$0.20 & 77.14\scriptsize$\pm$0.06 \\
RKD \cite{park2019relational} & 79.00\scriptsize$\pm$0.32 & 79.62\scriptsize$\pm$0.15 & 75.65\scriptsize$\pm$0.44\\
CRD \cite{tian2019contrastive} & 79.86\scriptsize$\pm$0.08 & 80.02\scriptsize$\pm$0.14 & 76.42\scriptsize$\pm$0.07 \\
\bottomrule
Teacher & \multicolumn{3}{c}{Teacher Free}  \\
Student & ResNet18 & ResNet34 & VGG16   \\
\midrule
Ours (4$\times$4) & \textbf{80.53\scriptsize$\pm$0.16} & \textbf{81.61\scriptsize$\pm$0.24} & \textbf{77.66\scriptsize$\pm$0.11} \\
\bottomrule
\end{tabular}
\end{center}
\end{table}

\begin{table}[htb!]
\centering
\renewcommand{\tabcolsep}{0.5mm} 
\caption{Accuracy (\%) on Fine-grained datasets. We report the averaged accuracy of three runs. The best results are indicated in \textbf{bold}.}
\label{test_accuracy_table_on_fine_grained}
\begin{center}
\begin{tabular}{l|cccc}
\toprule
\multicolumn{1}{c|}{Backbones} & \multicolumn{2}{c}{ResNet18} & \multicolumn{2}{c}{ResNet34}  \\   
\multicolumn{1}{c|}{Dataset} & CUB-200 & Stanford Dog & CUB-200 & Stanford Dog \\
\midrule
Hard Label & 57.76\scriptsize$\pm$0.51 & 66.14\scriptsize$\pm$0.17 & 59.79\scriptsize$\pm$0.60 & 68.90\scriptsize$\pm$0.34 \\
LS \cite{szegedy2016rethinking} & 60.31\scriptsize$\pm$0.43 & 66.81\scriptsize$\pm$0.19 & 63.57\scriptsize$\pm$0.49 & 69.85\scriptsize$\pm$0.26 \\
TF-KD\cite{yuan2020revisiting} & 58.28\scriptsize$\pm$0.28 & 66.23\scriptsize$\pm$0.23 & 61.88\scriptsize$\pm$0.38 & 69.78\scriptsize$\pm$0.58 \\

CS-KD\cite{yun2020regularizing} & 68.27\scriptsize$\pm$0.38 & 68.99\scriptsize$\pm$0.22 & 67.62\scriptsize$\pm$0.53 & 70.82\scriptsize$\pm$0.17 \\
\midrule
Ours (4$\times$4) & \makecell{\textbf{69.94}\scriptsize$\pm$\textbf{0.24} \\ (\textcolor{ForestGreen}{12.18$\uparrow$})} & \makecell{\textbf{70.94}\scriptsize$\pm$\textbf{0.07} \\ (\textcolor{ForestGreen}{4.80$\uparrow$})} & \makecell{\textbf{69.46}\scriptsize$\pm$\textbf{0.37} \\ (\textcolor{ForestGreen}{9.67$\uparrow$})} & \makecell{\textbf{71.75}\scriptsize$\pm$\textbf{0.13} \\ (\textcolor{ForestGreen}{2.85$\uparrow$})} \\
\bottomrule
\end{tabular}
\end{center}
\end{table}

\noindent\textbf{Fine-grained image classification: } Fine-grained image classification addresses the intricate challenge of distinguishing between visually similar subcategories within specific classes. Fine-grained datasets often exhibit high intra-class variability, where subtle difference in appearance, pose, or context exist among instances of the same class.  Consequently, these variations demand models capable of capturing fine details and making precise distinctions. In our framework, wherein two intra-class instances are encouraged to be consistent during training, fine-grained image classification becomes a crucial task to validate the superiority of the proposed method.

Table \ref{test_accuracy_table_on_fine_grained} presents the results of our method on fine-grained image classification with CUB-200-2011 and Standford Dogs datasets. Our method shows a significant improvement in performance on this task. This improvement can be attributed to the idea that our model achieves a balance between learning from both hard-to-learn and easy-to-learn samples through the self-distillation process, allowing it to capture fine details of the object within intra-class.

\subsection{Semantic Segmentation}
In the following two sections, we explore the transferability of the proposed method beyond image classification, extending our focus to the domains of semantic segmentation and object detection. We aim to extend the applicability of our approach by leveraging the pretrained network as a robust backbone for these additional tasks. This investigation examines the capacity of our method to generalize and transfer learned features to intricate tasks. 

\noindent\textbf{Implementation details: }
We use DeepLabV3+ \cite{chen2018encoder} as the segmentation network, and ResNet50 \cite{he2016deep} is used as the backbone network trained on the ImageNet and compare its performance with the network trained with hard labels.

We evaluate our approach on two different datasets, Pascal VOC2012 \cite{pascal-voc-2012} and Cityscapes \cite{cordts2016cityscapes}, as did the previous study \cite{chen2018encoder}. We set the input size to $513\times513$ and train the network for 50 epochs in the Pascal VOC2012, for 200 epochs in the Cityscapes, respectively. The initial learning rate is set to 0.007 in the Pascal VOC2012 and 0.01 in the Cityscapes with a batch size of $24$, respectively. We use the SGD optimizer with a momentum of 0.9 and a weight decay of 0.0005 for all datasets. Also, the poly learning rate is used in the learning rate scheduler with a power value of 0.9. We measure the mean Intersection over Union (mIoU) for evaluation.

\noindent\textbf{Results of semantic segmentation: }As illustrated in in Table \ref{test_accuracy_table_on_semantic_segmentation}, our method exhibits substantial improvements, with an increase of $2.79\%$ on VOC2012 and $2.17\%$ on Cityscapes. These results highlight the method's efficacy in transferring knowledge, showcasing a high level of generalizability in feature representation. The notable performance of our method can be attributed to the insight that our model captures not only the most discriminative parts but also comprehends the relevant aspects of the object, contributing to its enhanced segmentation capabilities.

\begin{table}[htb!]
\centering
\renewcommand{\tabcolsep}{4.5mm} 
\caption{Semantic segmentation results with mean Intersection of over Union (mIoU). `Hard Label' represents a result when a backbone network uses the ResNet50 model without distillation.}
\label{test_accuracy_table_on_semantic_segmentation}
\begin{center}
\begin{tabular}{@{} c *{3}{S} @{}}  
\toprule
\multirow{3}{*}{Method}  & \multicolumn{2}{c}{\makecell{Segmentation\\DLV3+ResNet50}} \\ 
\cmidrule(lr){2-3}\cmidrule(l){4-0}
 & {VOC 2012 val} & {Cityscapes val}   \\
\midrule
Hard Label & 72.46 &  67.09 \\
Ours (4$\times$4) & \textbf{75.25} & \textbf{69.26} \\
\bottomrule
\end{tabular}
\end{center}
\end{table}
\noindent\textbf{Coarse-grained class segmentation:} We provide further assessment into the results of semantic segmentation, with a particular emphasis on unstructured off-road scenes exemplified by the RUGD dataset \cite{wigness2019rugd}. Addressing perception in unstructured environments poses greater challenges due to the absence of clear boundaries and distinct features for many object classes. Therefore, tackling this task becomes crucial for evaluating the generalizability of feature representation. 

The classes in the RUGD dataset are organized into two levels of granularity: coarse-grained and fine-grained classes. Coarse-grained classes represent top-level categories, including smooth regions, rough regions, bumpy regions, forbidden regions, obstacles, and background. Conversely, fine-grained classes comprise subcategories within each coarse-grained class. Table \ref{class_category_RUGD} provides a summary of the class categories.

\begin{table}[htb!]
\centering
\caption{Class category on RUGD dataset \cite{wigness2019rugd}.}
\label{class_category_RUGD}
\begin{center}
\begin{tabular}{l|c}
\toprule
Coarse-class & Fine-class\\
\midrule
1. Smooth Region (SR) & concrete, asphalt \\
\midrule
2. Rough Region (RR) & gravel, grass, dirt, sand, mulch \\
\midrule
3. Bumpy Region (BR) & rock, rock-bed \\
\midrule
4. Forbidden Region (FR) & water \\
\midrule
5. Obstacle (Obs) & \makecell{tree, pole, vehicle, container, \\ building, log,
bicycle, person, \\ fence, bush, picnic-table, bridge} \\
\midrule
6. Background (Bac) & void, sky, sign \\
\bottomrule
\end{tabular}
\end{center}
\end{table}

In alignment with a previous study by Guan et al. \cite{guan2022ga}, this experiment involves conducting semantic segmentation using the coarse-grained class of the RUGD dataset. We used a distilled ResNet50 network trained on ImageNet as the backbone and a PSPHead decoder \cite{zhao2017pyramid}. The PSPHead is a head module in the Encoder-Decoder architecture that aims to capture global context information by aggregating multi-scale feature maps. It consists of a Pyramid Pooling module and several convolutional layers. The Pyramid Pooling module pools features at different scales and then upsamples them to the same size before concatenating them. This way, PSPHead can capture contextual information at different scales and produce a more accurate segmentation output \cite{zhao2017pyramid}. Additionally, we added a high-performing segmentation decoder head, such as FCN \cite{long2015fully}, in parallel to PSPHead.

\begin{table*}[htb!]
\caption{Comparison of our method with state-of-the-art techniques on RUGD: All reported results are from \cite{guan2022ga}. We compared the performance of our method with a transformer-based method (marked with *) and other approaches in terms of IoU and Acc. We highlight our results that surpass the baseline `Hard Label', which utilizes the ResNet50 backbone without distillation. `Hard Label' and `Ours' are the results obtained by combining the pretrained ResNet50 backbone with PSPHead + FCN decoder. \textbf{Bold} is the best}
\label{tab:coarse_segmentation}
\centering

\begin{center}
\scalebox{0.95}{

\begin{tabular}{l|cccccc|cc}
\toprule
  Methods (IoU) & SR & RR & BR & FR & Obs & Bac & mIoU & aAcc \\
\midrule
Hard Label & 72.11 & 92.78 & 88.57 & 83.71 & \textbf{91.73} & 76.33 & 84.21 & 94.90  \\
PSPNet~\cite{zhao2017pyramid} & 48.62 & 88.92 & 69.45 & 29.07 & 87.98 & 78.29 & 67.06 & 92.85 \\
DeepLabv3+~\cite{chen2018encoder} & 5.86 & 84.99 & 50.40 & 25.04 & 87.50 & 81.47 & 55.88 & 91.51 \\
DANet~\cite{fu2019dual} & 2.26 & 81.47 & 8.69 & 15.00 & 82.54 & 74.86 & 44.14 & 88.81 \\
OCRNet~\cite{yuan2020object} & 66.29 & 89.47 & 76.15 & 59.14 & 88.77 & \textbf{79.17} & 76.50 & 93.46 \\
PSANet~\cite{zhao2018psanet} & 34.92 & 87.70 & 35.64 & 8.66 & 86.95 & 78.97 & 55.47 & 92.13 \\
BiseNetv2\cite{yu2021bisenet} & 24.27 & 89.99 & \textbf{89.99} & 83.31 & 90.93 & 75.29 & 75.10 & 93.40 \\
CGNet~\cite{wu2020cgnet} & 40.84 & 90.39 & 85.67 & 76.21 & 89.75 & 74.48 & 76.22 & 93.29 \\
FastSCNN~\cite{poudel2019fast} & 83.03 & 92.82 & 87.69 & 81.05 & 90.94 & 75.11 & 85.11 & 94.77 \\
FastFCN~\cite{wu2019fastfcn} & 26.27 & 89.85 & 85.95 & 84.13 & 91.23 & 75.63 & 75.51 & 93.46 \\
*SETR~\cite{zheng2021rethinking} & 89.77 & 92.46 & 84.58 & 70.33 & 89.55 & 70.47 & 82.86 & 94.09 \\
*DPT~\cite{ranftl2021vision} & 1.04 & 81.23 & 22.98 & 25.84 & 89.18 & 74.50 & 49.13 & 88.77 \\
*Segformer\cite{xie2021segformer} & 93.26 & 93.16 & 87.56 & 77.31 & 91.20 & 78.50 & 86.83 & 95.17 \\
*Segmenter~\cite{strudel2021segmenter} & 90.39 & 91.17 & 83.96 & 65.43 & 87.80 & 68.17 & 81.15 & 93.22 \\
*GA-Nav-r32~\cite{guan2022ga} & 92.76 & 93.28 & 87.44 & 79.90 & 89.55 & 66.46 & 84.90 & 94.24 \\
\midrule
Ours (4$\times$4) & \textbf{94.22} & \textbf{94.01} & 89.22 & \textbf{84.83} & 91.61 & 75.87 & \textbf{88.29} & \textbf{95.42}  \\
\bottomrule
\end{tabular}
}
\end{center}
\end{table*}

\begin{figure*}[htb!]
  \begin{center}
    \includegraphics[width=0.85\linewidth]{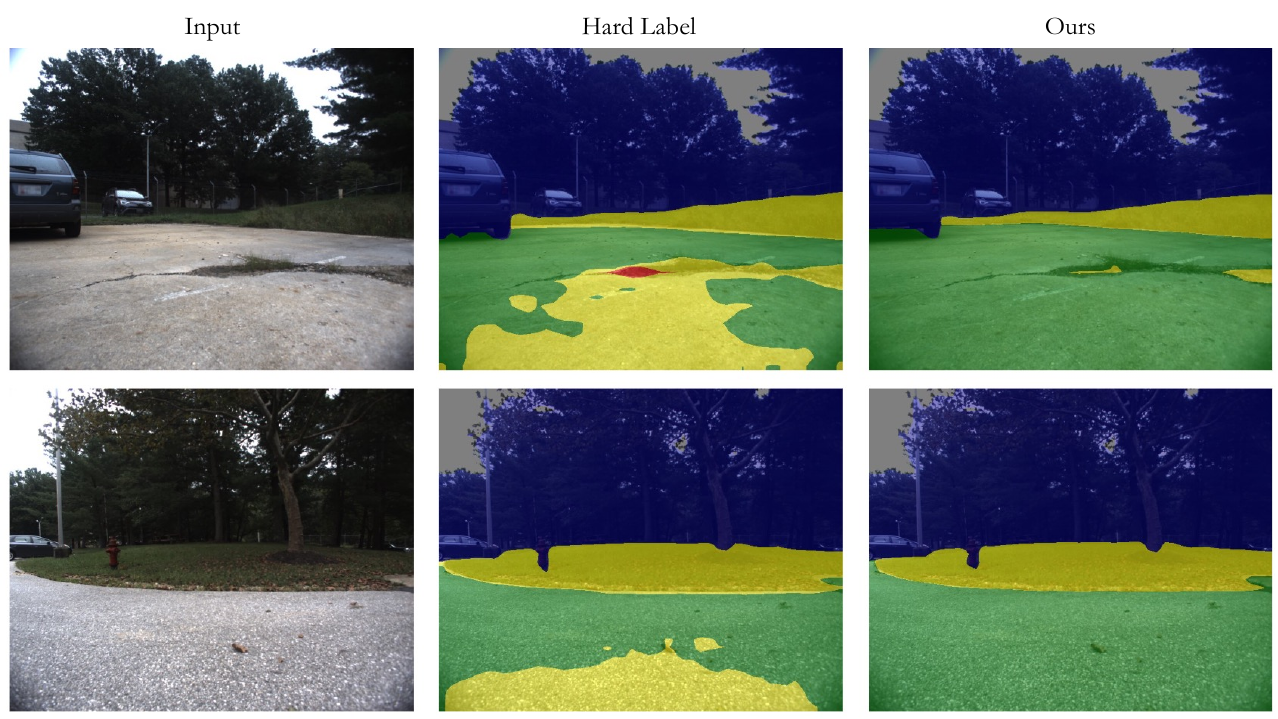}
\caption{Semantic segmentation results obtained from the RUGD dataset. The baseline method, utilizing Hard Labels, struggles to accurately identify the smooth region (SR, i.e., asphalt) and the rough region (RR, i.e., grass). In contrast, our approach demonstrates a notable advantage by providing clearer boundaries between each class.}
\label{fig:Seg_Results}
  \end{center}
\end{figure*}

As seen in Table \ref{tab:coarse_segmentation}, our approach achieves the best performance on a coarse-grained segmentation of the RUGD dataset. As compared to the baseline (Hard Label), our method presents the performance improvement (mIoU) by $4.08\%$. In addition to the quantitative results, Figure \ref{fig:Seg_Results} shows our qualitative results, which demonstrate our method's superior ability to segment concrete regions compared to the baseline approach. This is further supported by the quantitative results achieved: $72.11\%$ for the baseline versus $94.22\%$ for our method on SR (smooth region) class.

\noindent\textbf{Training details: } For `Hard Label' and `Ours' models, optimization is done using a stochastic gradient descent optimizer with a learning rate of 0.01 and a decay of 0.0005. We adopt the polynomial learning rate policy with a power of 0.9. The total iteration is 176K iterations. Also, we use a batch size of 8 and crop size of $300\times 375$ (the resolution of the original image is $688 \times 550$).

\subsection{Objection Detection}
\noindent\textbf{Implementation details: }For the object detection experiment, we use Single Shot Detection \cite{liu2016ssd} as the detector with the VOC2012 and VOC2007 datasets, where the union VOC2007 and VOC2012 are used for training and evaluating the model on the VOC2007 test set \cite{pascal-voc-2007}. The ResNet50 \cite{he2016deep} is used as the backbone network trained on the ImageNet. We set the input size to $300\times300$ and train the network for 250 epochs with a batch size of 32. We use the SGD optimizer with a momentum of 0.9 and a weight decay of 0.0005, and the learning rate warms up from 0 to 0.001 until 2 epochs and decays after 150 epochs and 200 epochs by multiplier 0.1. For the testing set, we use a confidence threshold of 0.01 and a non-maximum suppression overlap of 0.01. We assess performance using the mean Average Precision (mAP).

\begin{figure*}[htb!]
	\centering
	\includegraphics[width=0.99\linewidth]{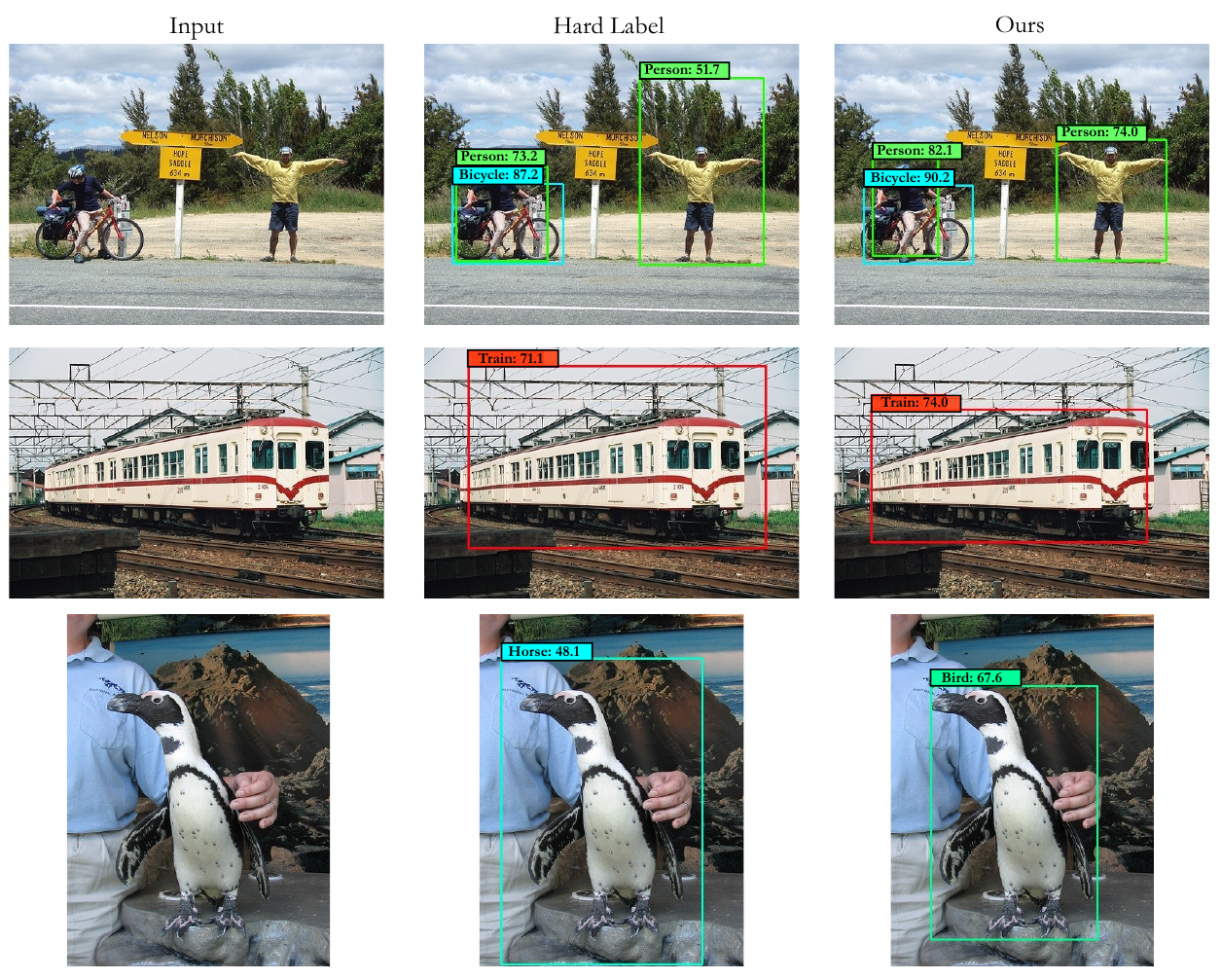}
	\caption{Three instances showcasing detection results of the baseline and the proposed method with SSD+ResNet50 on the VOC2007 test set. Notably, in the second input, the baseline model achieves an AP score of 71.1, while ours achieves a higher score of 74.0. Furthermore, in the case of the bird class, the baseline fails to predict the correct class, whereas our method accurately predicts it, providing a more compact bounding box.}\label{detection_qual}
\end{figure*}

\noindent\textbf{Results of object detection: }Table \ref{object_detection} shows the averaged precision (AP) over all classes. Outperforming the baseline for 14 out of 20 classes, our method demonstrates its efficacy in object detection. Additionally, qualitative results further support our observations in Figure \ref{detection_qual}. These results not only highlight our method's success in object detection but also substantiate our claim that it effectively improves feature representation.

\begin{table*}[htb!]
\renewcommand{\tabcolsep}{1.2mm} 
\centering
\caption{Object detection results with AP scores over all classes on on PASCAL VOC 2007 test set \cite{pascal-voc-2007} abd mean Average Precision (mAP). Here, `Hard Label' represents a result when a backbone network uses the ResNet50 model without distillation. \textbf{Bold} is the best.}
\label{object_detection}
\begin{center}
\scalebox{0.94}{\begin{tabular}{l|cccccccccc|c}
\toprule
 Method & Aeroplane & Bicycle & Bird & Boat & Bottle & Bus & Car & Cat & Chair & Cow  & mAP \\
 \midrule
Hard Label & 77.71 & 82.93 & 75.01 & 66.07 & 44.90  & \textbf{85.97} & 84.37 & \textbf{88.81} & 59.86 & 79.81 &  \\ 
Ours (4$\times$4) & \textbf{78.01} & \textbf{86.18} & \textbf{77.50} & \textbf{67.01} & \textbf{46.41}  & 85.26 & \textbf{85.54} & 87.44 & \textbf{63.03} & \textbf{83.06} & \\ \cmidrule{1-11}
& Dining Table & Dog & Horse & Motor Bike  & Person & Potted Plant & Sheep & Sofa & Train & TV Monitor & \\ \cmidrule{1-12}
Hard Label  & 76.44   & \textbf{86.58} & 86.08 & \textbf{84.10} & 78.28  & 49.30 & 75.22 & \textbf{80.32} & 85.25 & \textbf{75.70} & 76.13\\
Ours (4$\times$4) & \textbf{76.56} & 86.22 & \textbf{86.52} & 83.93 & \textbf{79.41}  & \textbf{53.61} & \textbf{78.18} & 79.59  & \textbf{87.01} & 75.31 & \textbf{77.29}\\
\bottomrule
\end{tabular}}
\end{center}
\end{table*}

\subsection{Ablation analysis}\label{further_analysis}
This section presents a comprehensive analysis of the proposed method, investigating its effectiveness across diverse dimensions, including reliability and robustness evaluations under corruptions and adversarial attacks \cite{yuan2019adversarial}, along with an examination of predictive distributions. Additionally, we provide comparative results with other data augmentation techniques. Through this extensive evaluation, we rigorously assess our models across an array of scenarios and challenges.

\subsubsection{Comparison with Other Data Augmentations} \label{analysis_aug}
We first present the comparative results with other augmentations such as CutMix \cite{yun2019cutmix}, CutOut \cite{devries2017improved}, and MixUp \cite{zhang2017mixup}. Note, as shown in accuracy result for (+Swap) of Table \ref{test_data_augmentation}, the swap itself has a negligible effect. Therefore, this swap module is not solely intended to diversify the training data when used alone; rather, a more critical role is to enable the interplay between `easy-to-learn' and `hard-to-learn' samples within our instance-to-instance framework. 

We assess the safety of ResNet50 on CIFAR100 using multiple metrics, including area under the risk-coverage curve (AURC) \cite{geifman2018bias}, false positive rate (FPR), expected calibration error (ECE) \cite{naeini2015obtaining}, and Brier score (BS) \cite{gneiting2007strictly}. AURC is to measure the area under the curve generated by plotting the risk according to coverage. ECE is a metric to measure calibration, implying the reliability of the model. BS is an important calibration metric to measure difference between the predicted probability and the actual outcome.
Furthermore, the model is evaluated against white-box attacks using two common attacks, FGSM and I-FGSM \cite{yuan2019adversarial}, with two perturbation values, 0.001 and 0.01. The corruption test is conducted using CIFAR100-C across ten corruptions, and the average accuracy is reported. As seen in Table \ref{test_data_augmentation}, our self-distillation roundly improves the performance on the safety axes while boosting test accuracy. Furthermore, in the next subsection, we demonstrate that our approach can be further improved in accuracy by combining it with other augmentations.

\begin{table*}[h!]
\renewcommand{\tabcolsep}{1.2mm} 
\caption{On CIFAR100, our method generally improves over other mix-based augmentations on a broad range of safety metrics. Higher is better except for the calibration test. \textbf{Bold} is the best, and \textcolor{red}{red} is second best.}
\label{test_data_augmentation}
\begin{center}
\scalebox{0.95}{\begin{tabular}{l|c|cccc|cccc|ccc}
\toprule
\multirow{3}{*}{ResNet50} & \multirow{3}{*}{\makecell{Accuracy \\ ($\uparrow$, \%)}} & \multicolumn{4}{c|}{Calibration($\downarrow$)} & \multicolumn{4}{c|}{Adversaries($\uparrow$, \%)} & \multirow{3}{*}{\makecell{Corruption \\ ($\uparrow$, \%)}} \\   
& & \makecell{AURC \\ ($\times10^{3}$)} & \makecell{FPR \\ ($\times10^{2}$)} & \makecell{ECE \\ ($\times10^{2}$)} & \makecell{BS \\ ($\times10^{2}$)} & \makecell{FGSM\\(0.001)} & \makecell{FGSM\\(0.01)} & \makecell{I-FGSM\\(0.001)} & \makecell{I-FGSM\\(0.01)} \\
\midrule
Hard Label & 78.7\scriptsize$\pm$0.13 & 56.1\scriptsize$\pm$0.24 & 59.7\scriptsize$\pm$0.14 & 10.6\scriptsize$\pm$0.15 & 32.1\scriptsize$\pm$0.22 & 74.9\scriptsize$\pm$0.16 & 45.9\scriptsize$\pm$0.50 & 74.9\scriptsize$\pm$0.17 & 37.3\scriptsize$\pm$0.68 & 55.4\scriptsize$\pm$0.32 \\
+Swap & 78.8\scriptsize$\pm$0.17 & 53.6\scriptsize$\pm$1.23 & \textcolor{red}{58.1}\scriptsize$\pm$0.62 & 9.57\scriptsize$\pm$0.17 & 31.2\scriptsize$\pm$0.16 & 75.6\scriptsize$\pm$0.21 & 48.9\scriptsize$\pm$1.28 & 75.6\scriptsize$\pm$0.21 & 42.2\scriptsize$\pm$2.63 & 54.3\scriptsize$\pm$0.51\\
+CutMix & \textcolor{red}{81.0}\scriptsize$\pm$0.21 & \textcolor{red}{51.0}\scriptsize$\pm$3.41 & 60.8\scriptsize$\pm$0.76 & \textcolor{red}{6.80}\scriptsize$\pm$1.75 & \textcolor{red}{28.3}\scriptsize$\pm$0.14 & 75.6\scriptsize$\pm$0.25 & 45.1\scriptsize$\pm$0.42 & 75.5\scriptsize$\pm$0.27 & 31.5\scriptsize$\pm$0.43 & 57.1\scriptsize$\pm$0.62 \\
+CutOut & 79.5\scriptsize$\pm$0.06 & 51.3\scriptsize$\pm$0.15 & 60.8\scriptsize$\pm$0.32 & 8.35\scriptsize$\pm$0.31 & 30.3\scriptsize$\pm$0.11 & \textcolor{red}{76.2}\scriptsize$\pm$0.29 & \textcolor{red}{50.6}\scriptsize$\pm$0.02 & \textcolor{red}{75.8}\scriptsize$\pm$0.08 & \textcolor{red}{42.8}\scriptsize$\pm$0.00 & 51.1\scriptsize$\pm$0.37\\
+MixUp & 80.0\scriptsize$\pm$0.06 & 55.1\scriptsize$\pm$0.26 & 58.7\scriptsize$\pm$1.67 & \textbf{4.79}\scriptsize$\pm$0.10 & 28.5\scriptsize$\pm$0.05 & 74.7\scriptsize$\pm$0.02 & 46.6\scriptsize$\pm$0.49 & 74.5\scriptsize$\pm$0.01 & 32.9\scriptsize$\pm$0.28 & \textbf{60.1}\scriptsize$\pm$0.38\\
\midrule
Ours ($4\times4$) & \textbf{81.9}\scriptsize$\pm$0.15 & \textbf{49.3}\scriptsize$\pm$1.18 & \textbf{56.8}\scriptsize$\pm$0.72 & 7.57\scriptsize$\pm$0.67 & \textbf{27.5}\scriptsize$\pm$0.14 & \textbf{78.4}\scriptsize$\pm$0.07 & \textbf{57.5}\scriptsize$\pm$0.30 & \textbf{78.2}\scriptsize$\pm$0.07 & \textbf{49.7}\scriptsize$\pm$0.10 & \textcolor{red}{59.3}\scriptsize$\pm$0.18 \\
\bottomrule
\end{tabular}}
\end{center}
\end{table*}

\noindent\textbf{Comprehensive results on corruption test: } Here, we present the detailed results of the corruption test involving all 10 categories. As depicted in Table \ref{corruption_test}, our methods consistently rank among the first or second-best approaches. These results attribute our success to the distinctive characteristics of our patch swap augmentation, ensuring the preservation of object integrity in augmented images without compromising any part of the original images. This unique attribute sets our approach apart from techniques like CutMix and CutOut, which involve random cropping of regions that may result in the loss of significant information, rendering them more susceptible to image corruption. In contrast to CutMix and CutOut, MixUp preserves image integrity through pixel-wise mixing, showcasing robustness against a diverse range of corruptions.

\begin{table}[htb!]
\renewcommand{\tabcolsep}{0.7mm} 
\centering
  \caption{Test accuracies for each type of corruption. The listed corruptions correspond to Defocus Blur (DB), Gaussian Blur (GB), Motion Blur (MB), Glass Blur (GB), Snow (SN), Frost (FR), Fog (FO), Brightness (BR), Contrast (CO), and Elastic Transform (ET). \textbf{Bold} is the best, and \textcolor{red}{red} is second best.}\label{corruption_test}
  \begin{center}
  \begin{tabular}{l|ccccc}
    \toprule
    \textbf{Types} & Hard Label & +CutMix & +CutOut &  +Mixup & Ours\\
    \midrule
    DB & 39.1\scriptsize$\pm$0.28 & 37.0\scriptsize$\pm$0.41 & 43.9\scriptsize$\pm$0.25 & \textcolor{red}{35.1}\scriptsize$\pm$0.31 & \textbf{34.5}\scriptsize$\pm$0.29 \\
    GB & 78.0\scriptsize$\pm$0.35 & 78.2\scriptsize$\pm$1.14  & 79.7\scriptsize$\pm$0.57 & \textcolor{red}{78.1}\scriptsize$\pm$1.21 & \textbf{76.6}\scriptsize$\pm$0.01 \\
    MB & 44.9\scriptsize$\pm$0.03 & 42.2\scriptsize$\pm$0.23 & 49.4\scriptsize$\pm$0.27 & \textbf{38.5}\scriptsize$\pm$0.37 & \textcolor{red}{40.1}\scriptsize$\pm$0.66 \\
    GB & 45.5\scriptsize$\pm$0.55 & 43.8\scriptsize$\pm$0.95 & 51.8\scriptsize$\pm$0.07 &  \textcolor{red}{40.5}\scriptsize$\pm$0.33 & \textbf{40.5}\scriptsize$\pm$0.23  \\
    SN & 43.4\scriptsize$\pm$0.30  & 39.9\scriptsize$\pm$0.74 & 45.3\scriptsize$\pm$0.75 & \textbf{37.1}\scriptsize$\pm$0.38 & \textcolor{red}{39.0}\scriptsize$\pm$0.08 \\
    FR & 50.7\scriptsize$\pm$0.35 & 49.6\scriptsize$\pm$0.80 & 56.0\scriptsize$\pm$0.87 & \textbf{44.7}\scriptsize$\pm$0.11 & \textcolor{red}{47.7}\scriptsize$\pm$0.05\\
     FO & 35.6\scriptsize$\pm$0.64 & 33.8\scriptsize$\pm$0.28 & 41.4\scriptsize$\pm$0.28 & \textbf{30.2}\scriptsize$\pm$0.11  &  \textcolor{red}{30.9}\scriptsize$\pm$0.08 \\
     BR & 25.2\scriptsize$\pm$0.18 & 23.7\scriptsize$\pm$0.02 & 26.2\scriptsize$\pm$0.14 & \textcolor{red}{23.4}\scriptsize$\pm$0.09 & \textbf{22.6}\scriptsize$\pm$0.30 \\
     CO & 45.0\scriptsize$\pm$0.30 & 41.9\scriptsize$\pm$0.09  & 52.6\scriptsize$\pm$0.48 & \textbf{34.7}\scriptsize$\pm$0.78 &  \textcolor{red}{39.4}\scriptsize$\pm$0.12\\
     ET & 38.1\scriptsize$\pm$0.28 & 37.7\scriptsize$\pm$1.56 & 42.2\scriptsize$\pm$0.09 &  \textbf{34.8}\scriptsize$\pm$0.17 & \textcolor{red}{35.0}\scriptsize$\pm$0.01\\
     \midrule
     Avg & 55.4\scriptsize$\pm$0.32 & 57.1\scriptsize$\pm$0.62 & 51.1\scriptsize$\pm$0.37 &  \textbf{60.1}\scriptsize$\pm$0.38 & \textcolor{red}{59.3}\scriptsize$\pm$0.18\\     
    \bottomrule
  \end{tabular}
\end{center}
\end{table}

\noindent\textbf{Comprehensive results on adversarial attacks: } Table \ref{adversarial} presents additional results for various perturbations of $\epsilon$ concerning both FGSM and I-FGSM attacks. Our method consistently demonstrates robustness across all values of $\epsilon$ with substantial margins. As the value of $\epsilon$ increases, the performance degradation for other methods significantly intensifies. In contrast, our method is less susceptible to attacks and exhibits less performance degradation. This is prominently presented with 0.05 of $\epsilon$ for the I-FGSM attack. All baselines show less than 5$\%$ of classification accuracy, whereas ours show over 20$\%$. This outcome suggests that our method possesses superior capabilities to withstand attacks and provides robustness in model safety.

\begin{table}[htb!]
\renewcommand{\tabcolsep}{0.6mm} 
\centering
\caption{Classification accuracy (\%) against the white-box attack with various perturbations of $\epsilon$ for every pixel on CIFAR100. \textbf{Bold} is the best, and \textcolor{red}{red} is the second best.}\label{adversarial}
\begin{center}
\begin{tabular}{l|ccccc}
\toprule
 & \multicolumn{5}{c}{FGSM \cite{yuan2019adversarial}}  \\
\midrule
eps & Hard Label & +MixUp & +CutOut & +CutMix & Ours  \\
\midrule
0 & 78.7\scriptsize$\pm$0.13 & 80.0\scriptsize$\pm$0.06 & 79.5\scriptsize$\pm$0.06 & \textcolor{red}{81.0}\scriptsize$\pm$0.21 & \textbf{81.9}\scriptsize$\pm$0.15  \\
0.001 & 74.9\scriptsize$\pm$0.16 & 74.7\scriptsize$\pm$0.02 & \textcolor{red}{76.2}\scriptsize$\pm$0.29 & 75.6\scriptsize$\pm$0.25 & \textbf{78.4}\scriptsize$\pm$0.07  \\
0.003 & 66.4\scriptsize$\pm$0.48 & 65.0\scriptsize$\pm$0.24 & \textcolor{red}{69.6}\scriptsize$\pm$0.02 & 65.7\scriptsize$\pm$0.10 & \textbf{71.7}\scriptsize$\pm$0.11  \\
0.005 & 59.0\scriptsize$\pm$0.33 & 57.3\scriptsize$\pm$0.46 & \textcolor{red}{62.9}\scriptsize$\pm$0.12 & 57.3\scriptsize$\pm$0.01 & \textbf{66.1}\scriptsize$\pm$0.04 \\
0.01 & 45.9\scriptsize$\pm$0.50 & 46.6\scriptsize$\pm$0.49 & \textcolor{red}{50.6}\scriptsize$\pm$0.02 & 45.1\scriptsize$\pm$0.42 & \textbf{57.5}\scriptsize$\pm$0.30 \\
0.05 & 16.3\scriptsize$\pm$2.47 & \textcolor{red}{27.0}\scriptsize$\pm$4.21 & 22.8\scriptsize$\pm$0.81 & 16.5\scriptsize$\pm$0.01 & \textbf{39.3}\scriptsize$\pm$3.95 \\
\midrule
& \multicolumn{5}{c}{I-FGSM \cite{yuan2019adversarial}} \\
\midrule
eps & Hard Label & +MixUp & +CutOut & +CutMix & Ours \\
\midrule
0 & 78.7\scriptsize$\pm$0.13 & 80.0\scriptsize$\pm$0.06 & 79.5\scriptsize$\pm$0.06 & \textcolor{red}{81.0}\scriptsize$\pm$0.21 & \textbf{81.9}\scriptsize$\pm$0.15 \\
0.001 & 74.9\scriptsize$\pm$0.17 & 74.5\scriptsize$\pm$0.01 & \textcolor{red}{75.8}\scriptsize$\pm$0.08 & 75.5\scriptsize$\pm$0.27 & \textbf{78.2}\scriptsize$\pm$0.07 \\
0.003 & 65.7\scriptsize$\pm$0.43 & 62.6\scriptsize$\pm$0.40 & \textcolor{red}{68.3}\scriptsize$\pm$0.02 & 63.9\scriptsize$\pm$0.13 & \textbf{70.5}\scriptsize$\pm$0.01 \\
0.005 & 56.8\scriptsize$\pm$0.48 & 51.3\scriptsize$\pm$0.42 & \textcolor{red}{60.5}\scriptsize$\pm$0.11 & 52.4\scriptsize$\pm$0.16 & \textbf{63.1}\scriptsize$\pm$0.11 \\
0.01 & 37.3\scriptsize$\pm$0.68 & 32.9\scriptsize$\pm$0.28 & \textcolor{red}{42.8}\scriptsize$\pm$0.00 & 31.5\scriptsize$\pm$0.43 & \textbf{49.7}\scriptsize$\pm$0.10 \\
0.05 & 1.47\scriptsize$\pm$0.44 & \textcolor{red}{2.85}\scriptsize$\pm$0.14 & 2.37\scriptsize$\pm$0.16 & 1.38\scriptsize$\pm$0.29 & \textbf{20.3}\scriptsize$\pm$0.46 \\
\bottomrule
\end{tabular}
\end{center}
\end{table}

\noindent\textbf{Extension with other augmentations: }Recent studies have highlighted the complementary benefits of mixing-based augmentations in conjunction with knowledge distillation \cite{choi2022knowledge}, particularly in finely adjusting the smoothness of logits. Our investigation further emphasizes the critical role of smoothness in self-distillation when integrating mixing-based augmentations with our method, as illustrated in Table \ref{ours_cutmix}. Specifically, the `Baseline' refers to the model trained with each augmentation but without self-distillation (baseline without augmentations: 79.01$\pm$0.24). `No swap' corresponds to $p_{r}=0$, meaning that mixing-augmented inputs are used within an instance-to-instance distillation framework without patch-swap augmentation. Setting $p_{r}=1$ allows all input pairs in a batch to be swapped, while $p_{r}=0.1$ limits this to only 10\% of input pairs. Our findings shows that our self-distillation approach yields additional performance benefits with mixing-based augmentations by carefully tuning the smoothness through $p_{r}$. This improvement may be due to the inclusion of inter-class representation, which enriches the distillation process with diverse knowledge transferred \cite{gou2021knowledge}. However, as observed with $p_{r}=1$, an excessive increase in swapped inputs can compromise image integrity, potentially leading to degraded performance across all methods. Further analysis, consistent with findings from \cite{choi2022knowledge}, indicates that generating an excessive number of augmented images may overly smooth logits, thereby adversely impacting overall performance.

\begin{table}[htb!]
\renewcommand{\tabcolsep}{1.3mm} 
\centering
\caption{Classification results (\%) obtained using other augmentations with varying $p_r$ values on CIFAR100 with ResNet50. Note that the performance, achieved without employing given augmentations, yields an accuracy of 79.01$\pm$0.24.}
\label{ours_cutmix}
\begin{center}
\begin{tabular}{cccccccc}
\toprule
$p_r$ & No swap & $0.1$ & $0.3$ & $0.5$ & $0.8$ & $1.0$ & Baseline\\
\midrule
\makecell{Ours \\ +CutMix} & \makecell{82.89\\\scriptsize$\pm$0.07} & \makecell{\textbf{83.15}\\ \scriptsize$\pm$0.28} & \makecell{82.88\\ \scriptsize$\pm$0.21} & \makecell{82.96\\\scriptsize$\pm$0.54} & \makecell{82.46\\\scriptsize$\pm$0.51} & \makecell{81.57\\\scriptsize$\pm$0.38} & \makecell{81.00\\\scriptsize$\pm$0.21}\\
\midrule
\makecell{Ours \\ +CutOut} & \makecell{79.78\\\scriptsize$\pm$0.26} & \makecell{80.23\\\scriptsize$\pm$0.16} & \makecell{\textbf{81.66}\\\scriptsize$\pm$0.09} & \makecell{81.41\\\scriptsize$\pm$0.13} & \makecell{80.27\\\scriptsize$\pm$0.04} & \makecell{78.17\\\scriptsize$\pm$0.37} & \makecell{79.53\\\scriptsize$\pm$0.06}\\
\midrule
\makecell{Ours \\ +MixUp} & \makecell{80.22\\\scriptsize$\pm$0.36} & \makecell{81.74\\\scriptsize$\pm$0.17} & \makecell{81.81\\\scriptsize$\pm$0.03} & \makecell{\textbf{82.05}\\\scriptsize$\pm$0.15} & \makecell{81.69\\\scriptsize$\pm$0.40} & \makecell{80.09\\\scriptsize$\pm$0.11} & \makecell{80.00\\\scriptsize$\pm$0.06}\\
\bottomrule
\end{tabular}
\end{center}
\end{table}

\subsubsection{Tolerance to Noisy Labels}
This section investigates the regularization capability of our method in the presence of noisy data. We introduce label noise by flipping the labels of a given proportion of training samples, following the approach outlined in \cite{wang2019symmetric}. The classification results for the ResNet18 model on CIFAR-100, with noise rates set to 20\%, 40\%, and 80\%, are presented in Table \ref{tolerance_to_noisy}. Notably, our method achieves the highest test accuracy, showcasing its effectiveness in mitigating overfitting to noisy samples.

\begin{table}[h!]
\caption{Classification results (\%) under different noise rates for training set.}
\label{tolerance_to_noisy}
 \begin{center}
\begin{tabular}{lccc}
\toprule
Noise rate & Hard Label & LS \cite{szegedy2016rethinking}& Ours \\
\midrule
0\% & 77.92\scriptsize$\pm$0.10 & 78.78\scriptsize$\pm$0.15 & \textbf{80.27\scriptsize$\pm$0.13} \\
20\% & 68.72\scriptsize$\pm$0.05 & 69.84\scriptsize$\pm$0.16 & \textbf{71.84\scriptsize$\pm$0.19} \\
40\% & 63.47\scriptsize$\pm$0.28 & 64.46\scriptsize$\pm$0.12 & \textbf{65.57\scriptsize$\pm$0.34}\\
80\% & 32.81\scriptsize$\pm$0.38 & 34.46\scriptsize$\pm$0.42 & \textbf{38.36\scriptsize$\pm$0.32}\\
\bottomrule
\end{tabular}
 \end{center}
\end{table}

\subsubsection{Enhanced Evaluation of Prediction Quality}\label{qual_pred}
In this section, we explore an advanced analysis of prediction quality to gain deeper insights into the effectiveness of the proposed method. By assessing the performance through predictive distribution and activation map from model outputs, we aim to offer a qualitative evaluation that helps us understand the relationships among predicted classes and highlights intriguing area within the spatial regions of the output. This analysis contributes to a more comprehensive understanding of the model's predictive capabilities and the meaningful features it captures.

\begin{figure}[htb!]
	\centering
	\includegraphics[width=0.6\linewidth]{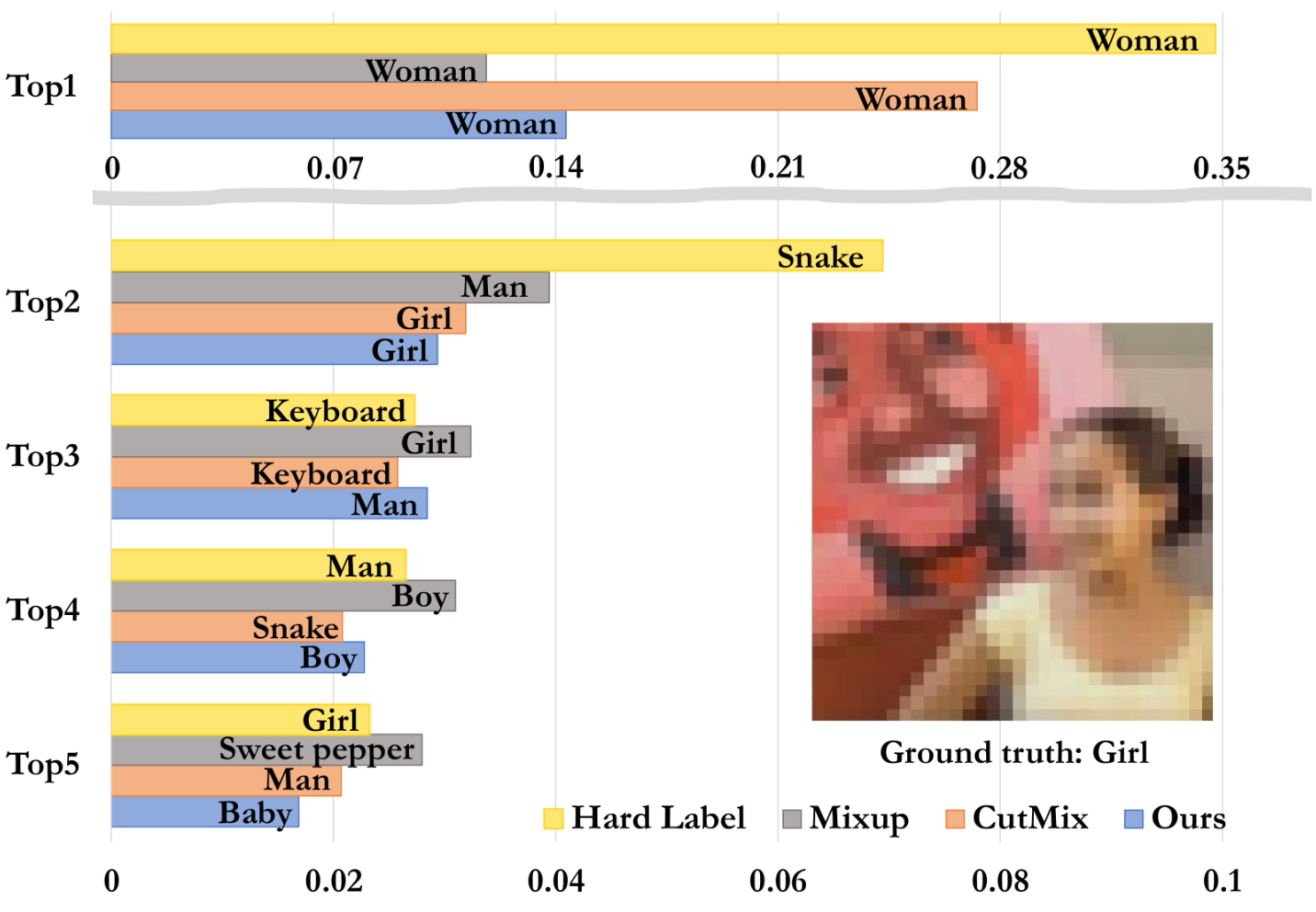}
	\caption{Predictive distributions on a misclassified example from CIFAR100. Our method assigns top-5 probability classes within the people class.}\label{predict_prob}
\end{figure}

\noindent\textbf{Predictive distributions: }
We first present the predictive distributions on CIFAR100 in Figure \ref{predict_prob}. We visualize the probability of a misclassified example with top-5 scores. As shown in Figure \ref{predict_prob}, for the attached image from CIFAR100 with ground truth labels of a girl, our method makes two desirable effects: better quality of predictive distribution and preventing overconfident predictions. Specifically, while baselines output the top-5 assigned classes including semantically unrelated classes (e.g., keyboard, snake, and sweet pepper), all top-5 assigned classes by our method belong to the people class (woman, girl, man, boy, and baby). 

\begin{figure}[htb!]
	\centering
	\includegraphics[width=0.73\linewidth]{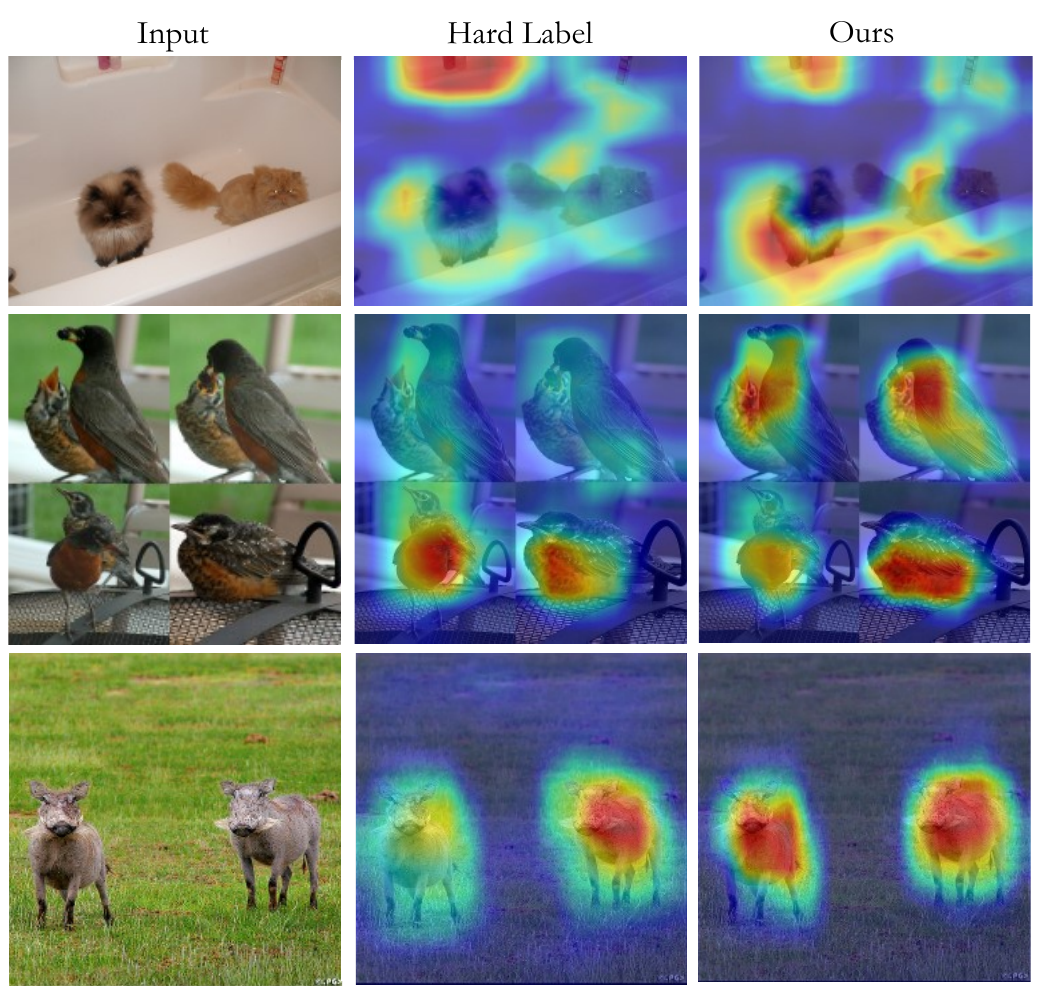}
	\caption{The activation maps generated by the Grad-CAM of the three examples of ImageNet validation set.}\label{grad-cam}
\end{figure}

\noindent\textbf{Activation map visualizations: } Visualization techniques like Grad-CAM \cite{selvaraju2017grad} play a vital role in helping us make sense of the model's outcomes and grasp the visual clues driving accurate predictions. This offers valuable insights into how the network reacts to the given inputs to make predictions. In the visualization, the red areas highlight important features that significantly contribute to the classification, while the blue areas represent less influential features. As shown in Figure \ref{grad-cam}, our approach pays more attention to the objects related to the target class, indicating a stronger performance in classification.

\subsubsection{Effects of Hyperparameters}\label{effect_hyperparameter} 
Throughout the preceding sections, we have highlighted the simplicity and applicability of our method in various computer vision tasks. In this section, we turn our attention to the exploration of hyperparameter settings, investigating their impact on the performance of our approach. 

\begin{figure}[htb!]
	\centering
	\includegraphics[width=0.7\linewidth]{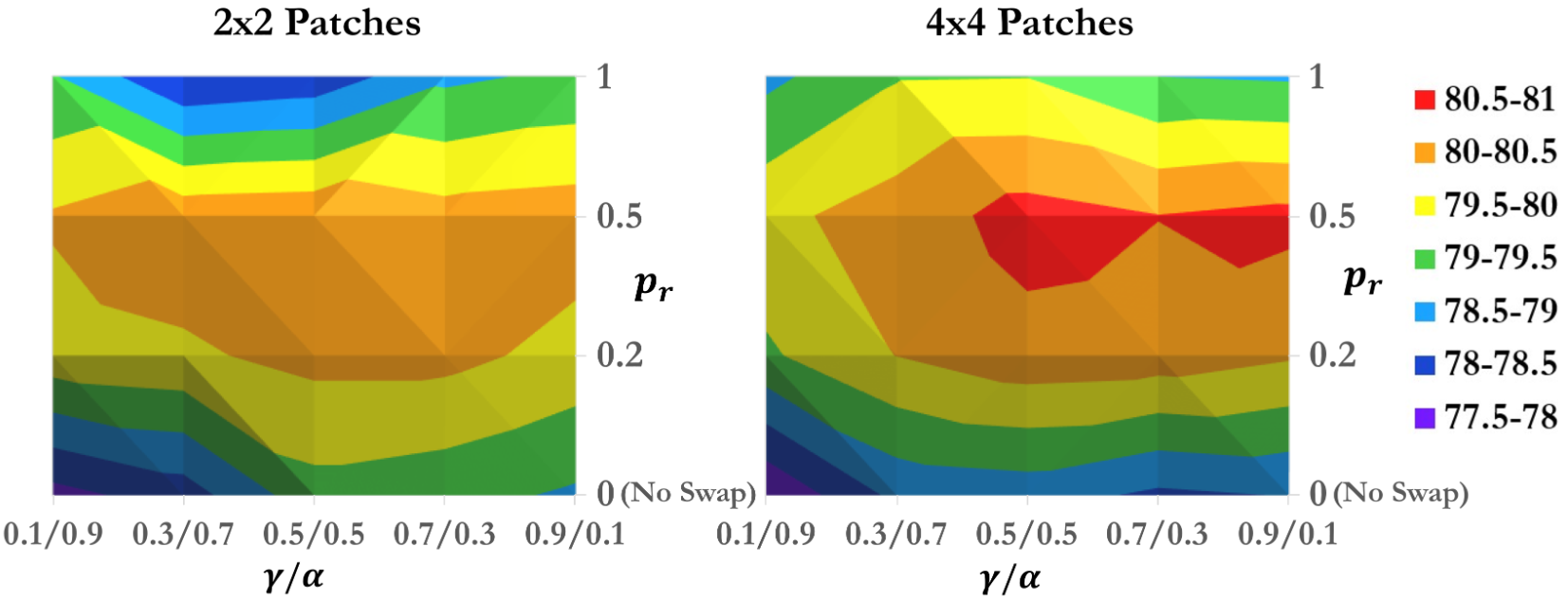}
	\caption{Best Top-1 accuracy for all combinations in the hyperparameters with ResNet18 on CIFAR100.}\label{hyper_effect}
\end{figure}

Figure \ref{hyper_effect} shows the sensitivity of our method to all hyperparameters used in our method, $\gamma$, $\alpha$, $p_{r}$, and a number of patches. Compared to the results without the swap ($p_{r}=0$), the swap with $p_{r}=0.5$ generally leads to the best performance, covering the high-accuracy region in the heatmaps. We can see how significantly the devised method affects distillation performance. Also, compared to 2$\times$2, 4$\times$4 performs better in most of the cases with different $\gamma$ and $\alpha$. This implies leveraging a smaller sized patch may help model in understanding small and partial knowledge. We discuss more in Section \ref{sec:discussion}.

\noindent\textbf{Gradual $p_r$ increment enhances performance: }When dealing with relatively small-scale datasets like CIFAR100, we have empirically observed that training models with `easy-to-learn' samples at the beginning can contribute to performance improvements. This advancement is achieved by progressively increasing the value of $p_{r}$ from 0.1 to 0.5 as the training process proceeds. We implement this approach according to the following protocol: for a training duration of 240 epochs, if the epoch is $\leq$ 30, then $p_r=0.1$; if the epoch is $\leq$ 80, then $p_r=0.2$; if the epoch is $\leq$ 120, then $p_r=0.3$; otherwise, $p_r=0.5$. Consequently, we systematically increase the likelihood of the given inputs being swapped as the training progresses. The impact of incremental $p_r$ in training process is demonstrated in Table \ref{varyingP}. Compared to using constant $p_r$ ($\thickbar{p_r}$), training with incremental $p_r$ ($\widehat{p_r}$) shows better performance.

\begin{table*}[htb!]
\renewcommand{\tabcolsep}{1.0mm} 
\caption{CIFAR100 test accuracy (\%) for networks trained with constant $p_r$ (represented as $\thickbar{p_r}$) and varying $p_r$ (denoted as $\widehat{p_r}$). The best results are highlighted in \textbf{bold}. The reported top-1 accuracy represents the mean of three runs. The performance gap with the Hard Label is shown in green.}
\label{varyingP}
\begin{center}
\scalebox{0.95}{\begin{tabular}{l|cccccccc}
\toprule
Method & ResNet18 & ResNet34 & ResNet50 & ResNet101 & VGG13 & VGG16 & ShuffleNetV1 & ShuffleNetV2 \\
\midrule
Hard Label & 77.92\scriptsize$\pm$0.10 & 79.05\scriptsize$\pm$0.09 & 79.01\scriptsize$\pm$0.24 & 79.29\scriptsize$\pm$0.21 & 74.62\scriptsize$\pm$0.23 & 74.40\scriptsize$\pm$0.03 & 71.26\scriptsize$\pm$0.11 & 72.62\scriptsize$\pm$0.18 \\
DLB\cite{shen2022self} & 80.25\scriptsize$\pm$0.15 & 80.71\scriptsize$\pm$0.12 & 81.35\scriptsize$\pm$0.15 & 81.83\scriptsize$\pm$0.06 & 77.08\scriptsize$\pm$0.38 & 76.58\scriptsize$\pm$0.19 & 73.80\scriptsize$\pm$0.31 & 74.30\scriptsize$\pm$0.25 \\
\midrule
Ours ($\thickbar{p_r}$) & \makecell{80.27\scriptsize$\pm$0.13 \\ (\textcolor{ForestGreen}{2.35$\uparrow$})} & \makecell{81.17\scriptsize$\pm$0.21 \\ (\textcolor{ForestGreen}{2.12$\uparrow$})} & \makecell{81.96\scriptsize$\pm$0.12 \\ (\textcolor{ForestGreen}{2.95$\uparrow$})} & \makecell{82.22\scriptsize$\pm$0.22 \\ (\textcolor{ForestGreen}{2.93$\uparrow$})} & \makecell{77.48\scriptsize$\pm$0.11 \\ (\textcolor{ForestGreen}{3.08$\uparrow$})} & \makecell{77.39\scriptsize$\pm$0.16 \\ (\textcolor{ForestGreen}{2.99$\uparrow$})} &  \makecell{74.21\scriptsize$\pm$0.07 \\ (\textcolor{ForestGreen}{2.95$\uparrow$})} & \makecell{75.51\scriptsize$\pm$0.09 \\ (\textcolor{ForestGreen}{2.89$\uparrow$})}  \\
Ours ($\widehat{p_r}$) & \makecell{\textbf{80.53}\scriptsize$\pm$\textbf{0.16} \\ (\textcolor{ForestGreen}{2.61$\uparrow$})} & \makecell{\textbf{81.61}\scriptsize$\pm$\textbf{0.24} \\ (\textcolor{ForestGreen}{2.56$\uparrow$})} & \makecell{\textbf{81.97}\scriptsize$\pm$\textbf{0.22} \\ (\textcolor{ForestGreen}{2.96$\uparrow$})} & \makecell{\textbf{82.71}\scriptsize$\pm$\textbf{0.12} \\ (\textcolor{ForestGreen}{3.42$\uparrow$})} & \makecell{\textbf{77.90}\scriptsize$\pm$\textbf{0.02} \\ (\textcolor{ForestGreen}{3.28$\uparrow$})} & \makecell{\textbf{77.66}\scriptsize$\pm$\textbf{0.11} \\ (\textcolor{ForestGreen}{3.26$\uparrow$})} &  \makecell{\textbf{74.32}\scriptsize$\pm$\textbf{0.29} \\ (\textcolor{ForestGreen}{3.06$\uparrow$})} & \makecell{\textbf{75.75}\scriptsize$\pm$\textbf{0.22} \\ (\textcolor{ForestGreen}{3.13$\uparrow$})}  \\
\bottomrule
\end{tabular}}
\end{center}
\end{table*}

We explain the observed empirical phenomenon by drawing inspiration from the principles of curriculum learning \cite{bengio2009curriculum}. At the beginning of training, the model is trying to learn the underlying patterns in the data, augmenting the data with `hard-to-learn' examples generated by high $p_{r}$ might introduce more complexity to the training process, making it harder for the model to find a good initial set of weights. As the training progresses and the model approaches convergence, introducing more challenging examples might help it fine-tune its parameters for better performance.

\section{Discussion} \label{sec:discussion}
For the proposed method, we introduced leveraging event difficulty manipulation for improved self-distillation.
`easy-to-learn' and `hard-to-learn' describe relative confidence levels of the two intra-class samples after patch swapping, rather than to imply absolute semantic difficulty. As shown in Figure \ref{overall_framework}, the distinction emerges naturally during training, one swapped image may retain more discriminative parts (e.g., an animal's head), producing high confidence, while the other may contain less discriminative features, producing lower confidence. However, this effect is not guaranteed in every case, since our approach does not explicitly select semantically important regions via attention maps. Instead, we deliberately trades precise region control for simplicity, computational efficiency, and general applicability by adopting randomized patch exchange.
To reduce risk of introducing misleading features, the swaps are constrained by a fixed patch size and occur only between semantically aligned (same-class) samples, which helps preserve label integrity. Also, to encourage robust feature learning, we explore dynamic augmentation strength in Table \ref{varyingP}, where we compare models trained with a fixed patch swap probability ($\bar{p}_r$) versus a progressively increasing schedule ($\hat{p}_r$). The progressive approach begins with low swap probability to allow the model to first learn strong and clean features from unaltered inputs. As training progresses, the increasing $p_r$ introduces greater difficulty through patch swaps, thereby fostering more dynamic and effective interplay between high and low confidence predictions. The improved performance of the progressive schedule suggest that modulating augmentation strength over time enhances joint optimization, acting as a form of curriculum learning: the model first stabilizes on core features before adapting to harder, perturbed variants.

In our study, we selected $2\times2$ and $4\times4$ patch sizes based on empirical validation, and these sizes were empirically found to be effective for maintaining useful variations in confidence levels between swapped pairs, while still preserving the overall structure necessary for reliable predictions. The effectiveness of patch size was described in Figure \ref{hyper_effect}.
For both 2$\times$2 and 4$\times$4 cases, the swap with $p_{r}=0.5$ generally leads to the better performance compared to without the swap ($p_{r}=0$). Also, across all experiments, utilizing 4$\times$4 patches outperform 2$\times$2 cases.
Using a significantly larger number of patches (i.e., lower resolution per patch) risks removing or distorting key discriminative parts of the input image, which can negatively impact the model's ability to learn core features from both samples. Utilizing much small sized patches not only provides hindrance to understand contexts in a patch, but also increases the complexity of patch swapping, which may be a cause of degradation.

In Table 15, we analyze the effect of progressively increasing the patch swap probability ($\hat{p}_r$) during training. This dynamic strategy begins with a low swap probability that allows the model to first learn from unperturbed samples and gradually increases the difficulty by introducing more patch-swapped inputs. This progressive learning schedule outperforms a fixed-probability baseline, highlighting the importance of preserving semantic integrity in the early training stage. From these observations, we infer that using a overly larger number of patches in training could reduce the chance of learning core features, thereby leading to degraded performance.

\section{Conclusion}\label{sec:conclusion}
In this study, we proposed a self-distillation framework based on intra-class patch swap augmentation, which consistently outperforms both the latest self-distillation methods and the conventional `Teacher-to-Student' distillation methods. A key factor behind this improvement is the combination of patch swap augmentation, which generates intra-class samples with varying prediction difficulty, and an instance-to-instance framework, which enables the model to effectively align their predictive distributions within a single network.
For intra-patch swap, by randomly swapping intra-class samples, two crucial attributes inject the predictive distribution: 1) constantly generate the event difficulty and 2) preserve semantically similar class relationships. We found that they have great benefits in distilling a single network. Our extensive experiments including image classification, semantic segmentation, and object detection demonstrate the effectiveness of the proposed method in a wide range of applications. This study reveals that the proposed method can produce superior self-generated knowledge with no powerful assistant networks in the classification tasks.

As can be seen in Section \ref{effect_hyperparameter}, the implementation of our method is straightforward, requiring minimal considerations for hyperparameters before model training such as patch size, loss weighting values ($\gamma$ and $\alpha$), and sampling probability ($p_{r}$). Our findings indicate that the efficacy of carefully tuning these hyperparameters in enchancing model performance. Synthesizing these observations reveals promising avenues for future research within the field of self-distillation. One intriguing direction is the exploration of automated approaches, perhaps leveraging meta-learning techniques \cite{finn2017model}. Such an approach could autonomously discover optimal hyperparameter configurations, eliminating the need for manual adjustments.
Also, optimal combinations for intra- and inter-class augmentations in a self-distillation could be explored to achieve further improvement.
This idea forms the basis of our future work.










\bibliographystyle{elsarticle-num} 
\bibliography{reference}

\end{document}